\RequirePackage{letltxmacro}
\LetLtxMacro{\LaTeXtextbf}{\textbf}
\documentclass{ieeeaccess}
\LetLtxMacro{\textbf}{\LaTeXtextbf}

\usepackage[utf8]{inputenc} 
\usepackage[T1]{fontenc}    
\usepackage{hyperref}       
\usepackage{url}            
\usepackage{booktabs}       
\usepackage{amsmath,amssymb,amsfonts}
\usepackage{nicefrac}       
\usepackage{microtype}      
\usepackage{multirow}
\usepackage{array}
\usepackage[usenames,dvipsnames]{xcolor}
\usepackage{balance}

\usepackage{graphicx}
\usepackage[ruled, linesnumbered]{algorithm2e}
\usepackage{algpseudocode}
\usepackage{subcaption}
\usepackage{tikz}
\usetikzlibrary{positioning}
\usetikzlibrary{calc}

\NewSpotColorSpace{PANTONE}
\AddSpotColor{PANTONE} {PANTONE3015C} {PANTONE\SpotSpace 3015\SpotSpace C} {1 0.3 0 0.2}
\SetPageColorSpace{PANTONE}%

\input{diagram_commands}

\usepackage{soul}
\newcommand{\revise}[1]{#1}

\newcommand{\deltheta}{\frac{\partial}{\partial \theta}}
\newcommand{\intA}[1]{\int_\mathcal{A} #1 ~da}

\newcommand{\df}{\mathcal{D}_f}

\usepackage[nogroupskip,acronyms,nopostdot,style=super,nonumberlist,toc]{glossaries}
\newacronym{drl}{DRL}{Deep Reinforcement Learning}
\newacronym{rl}{RL}{Reinforcement Learning}
\newacronym{mcmc}{MCMC}{Markov Chain Monte Carlo}
\newacronym{vi}{VI}{Variational Inference}
\newacronym{ac}{AC}{Actor-Critic}
\newacronym{kl}{KL}{Kullback-Leibler}
\newacronym{fkl}{FKL}{Forward Kullback-Leibler}
\newacronym{rkl}{RKL}{Reverse Kullback-Leibler}
\newacronym{js}{JS}{Jensen-Shannon}
\newacronym{kde}{KDE}{Kernel Density Estimation}
\newacronym{gan}{GAN}{Generative Adversarial Network}
\newacronym{pdf}{pdf}{probability density function}
\newacronym{sac}{SAC}{Soft Actor-Critic}
\newacronym{il}{IL}{Imitation Learning}
\newacronym{cb}{CB}{Contextual Bandit}
\newacronym{gail}{GAIL}{Generative Adversarial Imitation Learning}
\newacronym{irl}{IRL}{Inverse Reinforcement Learning}
\newacronym{cv}{CV}{computer vision}
\newacronym{nlp}{NLP}{natural language processing}
\newacronym{me}{ME}{maximum entropy}
\newacronym{cnn}{CNN}{convolutional neural network}
\newacronym{nf}{NF}{Normalizing Flow}
\def\BibTeX{{\rm B\kern-.05em{\sc i\kern-.025em b}\kern-.08em
    T\kern-.1667em\lower.7ex\hbox{E}\kern-.125emX}}

\begin{document}
\history{Date of publication xxxx 00, 0000, date of current version xxxx 00, 0000.}
\doi{10.1109/ACCESS.2024.3376739}

\title{Learning to Generate All Feasible Actions}
\author{\uppercase{Mirco Theile}\authorrefmark{1,2}, \IEEEmembership{Student Member, IEEE},
\uppercase{Daniele Bernardini}\authorrefmark{1,3}, \IEEEmembership{Member, IEEE}, \\\uppercase{Raphael Trumpp}\authorrefmark{1}, \IEEEmembership{Student Member, IEEE}, \uppercase{Cristina Piazza}\authorrefmark{3}, \IEEEmembership{Senior Member, IEEE}, \\\uppercase{Marco Caccamo}\authorrefmark{1}, \IEEEmembership{Fellow, IEEE}, \uppercase{Alberto L. Sangiovanni-Vincentelli}\authorrefmark{2},
\IEEEmembership{Fellow, IEEE}}

\address[1]{TUM School of Engineering and Design, Technical University of Munich (e-mail: \{mirco.theile,daniele.bernardini,raphael.trumpp,mcaccamo\}@tum.de)}
\address[2]{Dept. of Electrical Engineering and Computer Sciences, University of California, Berkeley (e-mail: alberto@berkeley.edu)}
\address[3]{TUM School of Computation, Information and Technology, Technical University of Munich (e-mail:
cristina.piazza@tum.de)}
\tfootnote{Marco Caccamo was supported by an Alexander von Humboldt Professorship endowed by the German Federal Ministry of Education and Research.}

\markboth
{Theile \headeretal: Learning to Generate All Feasible Actions}
{Theile \headeretal: Learning to Generate All Feasible Actions}

\corresp{Corresponding author: Mirco Theile (e-mail: mirco.theile@tum.de).}

\begin{abstract}
Modern cyber-physical systems are becoming increasingly complex to model, thus motivating data-driven techniques such as reinforcement learning (RL) to find appropriate control agents. However, most systems are subject to hard constraints such as safety or operational bounds. Typically, to learn to satisfy these constraints, the agent must violate them \revise{systematically}, which is \revise{computationally} prohibitive in most systems. Recent efforts aim to utilize feasibility models that assess whether a proposed action is feasible to avoid applying the agent's infeasible action proposals to the system. However, these efforts focus on guaranteeing constraint satisfaction rather than the agent's learning efficiency. 
\revise{To improve the learning process, we introduce} \textit{action mapping}\revise{, a novel approach that divides the learning process into two steps: first learn feasibility and subsequently,} the objective by mapping actions into the sets of feasible actions. 
This paper focuses on the feasibility part by \textit{learning to generate all feasible actions} through self-supervised querying of the feasibility model. We train the agent by formulating the problem as a distribution matching problem and deriving gradient estimators for different divergences. \revise{Through an illustrative example, a robotic path planning scenario, and a robotic grasping simulation, we demonstrate the agent's proficiency in generating actions across disconnected feasible action sets. By addressing the feasibility step, this paper makes it possible to focus future work on the objective part of action mapping,} paving the way for an RL framework that is both safe and efficient.
\end{abstract}


\titlepgskip=-21pt

\maketitle

\section{Introduction}

Cyber-physical systems are becoming increasingly complex, with applications ranging from autonomous vehicles in chaotic urban environments to robotic assistants for support in everyday tasks. Most of these applications require the development of complex control systems. Traditionally, these systems were modeled in detail, and control strategies were derived using model-based techniques. However, the increasing complexity of these systems limits the applicability of model-based techniques, thus making data-driven techniques appealing. While data-driven techniques such as reinforcement learning (RL) improved significantly in recent years, they still lack guarantees that they meet all system constraints, i.e., only providing \textit{feasible} control commands. 

A popular idea is deriving only the feasibility-relevant part of the system to ensure feasibility while using learning techniques to optimize the underlying objective. The feasibility model only delineates whether a suggested control command in a given situation is feasible, i.e., the control command does not violate any constraints and does not lead to a state from which a future constraint violation is inevitable. Given this feasibility model, the subsequent challenge is integrating it within a learning framework in which a policy aims to optimize an objective \revise{function} subject to feasibility constraints. The commonly applied techniques are \textit{action rejection}, \textit{resampling}, and \textit{action projection}.

Action rejection is a traditional approach, e.g., applied in the Simplex architecture \cite{bak2009system}, which can be summarized as follows. If the policy's proposed action is feasible according to the feasibility model, it is applied to the system. Otherwise, a backup policy is used, which generates a feasible action, usually independent of the objective. While this is the simplest method to implement, and the timing requirements are predictable, the drawback is that the policy needs to learn the feasibility model explicitly to avoid its action being rejected and replaced with the \revise{sub-optimal} backup action.

As a straightforward augmentation of the action rejection scheme, action resampling can be applied when training a stochastic policy. Instead of directly switching to the safe action, if the proposed action is infeasible, the policy can be resampled, and the newly generated action can be tested \cite{bharadhwaj2021conservative}. This process can be repeated until either a feasible action is proposed or a timeout is reached, at which point the safe action of the feasibility controller is applied to the system. While this method may decrease the rejection rate of the policy's actions, it adds computational costs. Additionally, most learning methods train agents that output a reparameterization of a single Gaussian. Resampling from this Gaussian may not offer a feasible action if it is too narrow or poorly aligned with the set of feasible actions. Moreover, the learning agent must still explicitly learn to avoid proposing infeasible actions.

A more nuanced method is action projection \cite{cheng2019end}, which replaces a proposed infeasible action with a feasible action closest to the proposed action. This projection is typically formulated as an optimization problem that must be solved online. The supposed advantage of this method over action rejection is that the replacement action is better than the safe action, which was derived independently of the objective. However, only because the projected action is \textit{close in the action space} does not mean it is also \textit{close in performance}. Additionally, the online optimization requirement may not be computationally feasible, especially for complex systems. From a learning perspective, the projection can either be penalized or ignored. If penalized, the agent again needs to learn explicitly to avoid infeasible actions, but it could receive guidance from the projection distance. If the agent does not penalize infeasible actions, the agent is not required to learn the feasibility model. However, the projection to the closest feasible action will map all infeasible actions to the borders of the feasible action sets. Learning algorithms that require action densities or policy gradients must be adapted to handle the resulting high action density on the borders.

In all three approaches, the learning agent that aims to find an optimum of the objective subject to the feasibility constraints is not aided by the feasibility model; it is solely made safe. \revise{The agent must still violate the constraints systematically during interactions with the environment, albeit without actually applying infeasible actions to the system, to learn to satisfy them in the future.} We \revise{introduce} a different approach that allows the learning agent to benefit explicitly from the model-based feasibility model. We call the approach \textit{action mapping}. The idea is to learn the feasibility and the objective consecutively. First, a \textit{feasibility policy} is trained to generate all feasible actions for a given state. Using this feasibility policy, an \textit{objective policy} can learn to choose the optimal action from the feasible ones, given an objective. Note that the optimization problem in the feasible actions could be solved with various methods, including, but not limited to, learning, which can all benefit from the guarantee of constraint satisfaction.

This methodology promises multiple potential advantages. First, the feasibility policy can be trained directly on the feasibility model, requiring no interactions with the environment. Afterward, the objective policy learns to choose among feasible actions, which could significantly reduce the number of interactions with the environment. \revise{The combined agent, i.e., feasibility plus objective policy, still needs to exhaustively violate constraints. However, it can learn constraint satisfaction offline from the feasibility model without interactions with the environment.} Second, the feasibility policy can be reused if multiple objectives are subject to the same constraints. Third, any knowledge of the environment that can be extracted from the feasibility model can potentially be utilized in the objective policy through parameter sharing between both policies. Lastly, once deployed, it requires precisely one pass of the feasibility policy and the objective policy per step if the feasibility policy has no support in the infeasible action space. 

Given these potential advantages, the pivotal question is: How do we train the feasibility policy? This paper endeavors to answer this very question. To \revise{this} end, we derive the objective of the feasibility policy as a distribution matching problem in which the target is a uniform distribution over the feasible action space. \revise{The uniform distribution is chosen since the feasibility policy is agnostic to the objective and should thus not be biased toward specific actions.} We further present a methodology for estimating the gradient of different divergence measures to train a feasibility policy toward the target distribution. To evaluate our proposed methodology, we \revise{perform three} experiments. The first is an illustrative example with an analytical and highly parallelizable feasibility function that \revise{shows} the input and output of the feasibility policy. \revise{The second example illustrates how the feasibility policy can learn to generate feasible trajectory segments for robotic path planning problems, providing a closer tie to reinforcement learning.} The \revise{third} experiment showcases a simple robotic grasping example where feasibility is defined as grasping poses that lead to a successful grasp. This experiment shows how a feasibility policy can be learned for systems without a feasibility model that can be efficiently parallelized. 

\revise{The contributions of this work are the following:}
\begin{itemize}
    \item \revise{Conceptualization of \textit{action mapping} as a framework for safe and efficient reinforcement learning;}
    \item \revise{Formulation of a distribution matching problem to train the feasibility policy towards generating all feasible actions;}
    \item \revise{Derivation of gradient estimators for different divergence measures utilizing kernel density estimates, resampling, and importance sampling;}
    \item \revise{Evaluation of the proposed approach in an illustrative 2D example, a qualitative example for spline-based path planning, and a quantitative planar robotic grasping example.}
\end{itemize}

The remainder of this paper is structured as follows. Section~\ref{sec:rel} discusses related work. Section~\ref{sec:opti} describes the action mapping motivation and the formulation as a distribution matching problem, followed by the gradient estimation in Section~\ref{sec:method}. Section~\ref{sec:illustrative} provides an illustrative example to visualize the feasibility policy \revise{and Section}~\ref{sec:splines} \revise{ provides an additional example that showcases how action mapping could be used in robotic path planning problems}. Sections~\ref{sec:setup} and~\ref{sec:results} introduce and discuss the robotic grasping experiments. 
\section{Related Work}
\label{sec:rel}

In discrete action spaces, the equivalent of action mapping is action masking, for which the feasibility of each action is evaluated, and the agent chooses the best action among the feasible ones. In~\cite{alshiekh2018safe}, the action masking concept is termed \textit{shielding}, in which the shield is based on linear temporal logic. The authors in~\cite{huang2022closer} investigate the consequences of action masking for policy gradient \revise{deep reinforcement learning (DRL)} algorithms. Applications in various domains show significant performance improvements, e.g., in autonomous driving~\cite{krasowski2020safe}, \revise{unmanned aerial vehicle (UAV)} path planning~\cite{theile2023learning}, \revise{and} vehicle routing~\cite{nazari2018reinforcement}.

For continuous action spaces, a straightforward masking approach is not yet available. As discussed before, the approaches can be grouped into \textit{action rejection}, \textit{resampling}~\cite{garcia2015comprehensive}, and \textit{action projection}~\cite{fisac2019general,li2018safe,dalal2018safe}. The safety model can be based on control barrier functions~\cite{ames2019control}, Lyapunov functions~\cite{sha2001using}, or variants thereof. Cheng et al. ~\cite{cheng2019end} use action projection and train a second model on the previous interventions to reduce the need for future interventions. Zhong et al. ~\cite{zhong2021safe} derive a \textit{safe-visor} that rejects infeasible actions proposed by the agent and replaces it with a safe action. 

The distribution matching problem is similar to posterior sampling, a long-standing problem in statistics. State-of-the-art methods in Bayesian statistics rely on \gls*{mcmc} algorithms \cite{hastings1970, gelfand1990}, eliminating the need to normalize the distribution, which is often an intractable problem \cite{kruschke2015}. \gls*{vi} relies instead on fitting the posterior with a family of parametric probability distributions that can be sampled \cite{Jordan1998, WainwrightJordan2008}. Neural samplers offer another alternative by approximating the posterior with a generative neural network \cite{nowozin2016, Hu2018}. 

\glspl*{nf} infer the \gls*{pdf} for each sample using invertible mappings \cite{Rezende2015, Tabak2013, Tabak2010}. While \glspl*{nf} do not require density estimates, they have been shown to require a prohibitive number of layers to effectively match a target distribution in more than one dimension \cite{kong2020expressive}. However, the depth of such models can lead to challenges like vanishing or exploding gradients, which are even exacerbated by the inherent conditioning difficulties of \gls*{nf}s \cite{pmlr-v139-koehler21a}. 

For robotic grasping, \revise{the authors in}~\cite{kalashnikov2018scalable} propose using \revise{DRL} to find optimal grasps through interaction with multiple real-world robots. If the goal is to find grasping poses explicitly to be used as the target of a classical controller, supervised learning techniques are often utilized \cite{kleeberger2020survey}. To support various downstream tasks, it would be necessary to find all feasible grasps. To this end, the action space is typically discretized, and grasping success is estimated for each discrete action through heat-maps. This can be learned \revise{using} supervised \cite{kumra2020antipodal, morrison2020learning} or self-supervised \cite{zeng2020tossingbot} \revise{methods}. \cite{zeng2020tossingbot} explicitly \revise{utilizes the} structure given by spatial equivariances. We aim to find a solution that needs neither discretization nor the use of the structure, as these requirements are specific to grasping and also restrict applicability to planar picking in carefully crafted environments.

\section{Optimization Problem}
\label{sec:opti}

\subsection{Action Mapping}

For a state space $\mathcal{S}$ and an action space $\mathcal{A}$, the feasibility model can be expressed through the function
\begin{equation}
    g:\mathcal{S}\times\mathcal{A}\to\mathbb{B},
\end{equation}
which delineates if a suggested action is feasible in a given state. Given $g$, the state-dependent set of feasible actions $\mathcal{A}_s^+\subseteq\mathcal{A}$ contains all actions that are feasible for the state $s$, i.e., all actions for which $g(s, a) = 1$. 

For action mapping, the feasibility policy is defined as
\begin{equation}
    \pi_\text{feasibility} : \mathcal{S} \times\mathcal{Z} \to \mathcal{A}_s^+.
\end{equation}
It learns a state-conditioned surjective map from a bounded latent space $\mathcal{Z}\subset \mathbb{R}^m$, with appropriate dimensionality $m$, into the set of feasible actions for that state. The latent space $\mathcal{Z}$ can be thought of as an infinite set of \textit{indices}. For each \textit{index}, the feasibility policy has to output a different feasible action.

Given the task specifics, an objective policy can be defined that learns the optimal latent value as 
\begin{equation}
    \pi_\text{objective} : \mathcal{S} \to \mathcal{Z} .
\end{equation}
This optimal \textit{index} in the latent space can then be mapped to a feasible action using $\pi_\text{feasiblity}$. Convolving the functions as $(\pi_\text{feasibility} \circ \pi_\text{objective}): \mathcal{S} \to \mathcal{A}_s^+$, yields the action mapping policy
\begin{equation}
    \pi(s) = \pi_\text{feasibility}(s, \pi_\text{objective}(s)).
\end{equation}

In this work, we derive how to train the feasibility policy. Since this work only concerns the feasibility policy, the subscript is dropped in the following.

\subsection{Feasibility Policy}
To train the feasibility policy $\pi_\theta$, we parameterize it with parameters $\theta$ and formulate a distribution matching problem. The goal is that $\pi_\theta$ maps every $z\in\mathcal{Z}$ to an $a\in\mathcal{A}_s^+$, without any bias toward any specific feasible actions. Therefore, by sampling uniformly in $\mathcal{Z}$, $\pi_\theta$ should generate a uniform distribution in $\mathcal{A}_s^+$.

When sampling uniformly in $\mathcal{Z}$, $\pi_\theta$ becomes a generator with a conditional \acrfull*{pdf} $q_\theta(a|s)$. The target distribution is the uniform distribution in the feasible action space given as
\begin{equation}
    \label{eq:scoredist}
    p(a|s) = \frac{g(s,a)}{\int_\mathcal{A}g(s,a')da'}.
\end{equation}
Given a divergence measure $\mathcal{D}$, the optimal parameters are the solution to the optimization problem
\begin{equation}
    \operatorname{argmin}_{\theta\in\Theta} \int_\mathcal{S}\mathcal{D}\big(p(\cdot|s) \,||\, q_\theta(\cdot|s)\big) ds,
\end{equation}
with $\Theta$ being the set of possible parameters. The following section details how to iteratively minimize the divergence.

\section{Methodology}\label{sec:method}
The following derives the gradient w.r.t. $\theta$ to iteratively minimize the divergence for a given state. For simplicity of notation, we omit the state and action dependence of $q_\theta$ and $p$.
\subsection{f-Divergence}
As the divergence measure, we choose the f-divergence, a generalization of the \gls*{kl} divergence (\cite{liese2006}). The f-divergence between two \glspl*{pdf} $p$ and $q_\theta$ has the form
\begin{equation}
    \label{eq:f-div}
    \df(p\,||\,q_\theta) = \intA{p ~f\left( \frac{q_\theta}{p} \right)},
\end{equation}
where $f: (0, \infty)\to \mathbb{R}$ is a convex function. Different choices of $f$ lead to well-known divergences as summarized in Table~\ref{tb:divergences}.
\begin{table}
\centering
\caption{Non-exhaustive list of f-divergences and the corresponding first derivative for gradient estimators.}
\begin{tabular}{l|c|c}
                        & $f(t)$ & $f'(t)$ \\ 
\hline
\acrshort*{js}           & $\frac{1}{2}\left[ (t+1)\log\left(\frac{2}{t+1}\right) + t~\log(t) \right]$ & $\frac{1}{2} \log\left(\frac{2t}{t+1}\right)$ \\
\acrshort*{fkl}        & $-\log(t)$ & $-\frac{1}{t}$ \\
\acrshort*{rkl}         & $t~\log(t)$ & $\log(t) + 1$ \\ \hline
\multicolumn{3}{p{230pt}}{\revise{The f-divergences are obtained by substituting the $f$ functions above in} \eqref{eq:f-div} \revise{and setting $t=q_\theta/p$. The conventions for $p$, $q$, FKL and RKL assume that $p$ is the target distribution, $q$ is the model, and the FKL divergence is $\int p\log(p/q)$.}}
\end{tabular}
\label{tb:divergences}
\end{table}
The gradients of the f-divergence w.r.t. $\theta$ can be estimated commuting the derivative with the integral (\cite{lecuyer1995sufficient}) and using the score function gradient estimator  (\cite{Kleijnen1996optimization}) as
\begin{align}
    \deltheta\df &= \deltheta\intA{p~f\left(\frac{q_\theta}{p}\right)} \nonumber \\
    &= \intA{p~f'\left(\frac{q_\theta}{p}\right)\frac{1}{p}\deltheta q_\theta} \nonumber \\
    &= \intA{q_\theta~f'\left(\frac{q_\theta}{p}\right)\deltheta \log q_\theta},
\end{align}
\revise{considering} that $p$ does not depend on $\theta$. Since $q_\theta$ is normalized to $1$ and thus $\partial_\theta\intA{q} = \intA{ q~\partial_\theta\log{q}} = 0$, a Lagrangian term $\lambda$ can be added to the gradient:
\begin{equation}
    \deltheta\df = \intA{q_\theta~\left(f'\left(\frac{q_\theta}{p}\right) +\lambda\right)\deltheta \log q_\theta}. \label{eq:dflambda}
\end{equation}
If the support of $q_\theta$ includes all of $\mathcal{A}$ the above formula \revise{in }\eqref{eq:dflambda} can be rewritten as the expectation on $q_\theta$ as
\begin{equation}
    \deltheta\df = \mathbb{E}_{q_\theta}\left[ \left(f'\left(\frac{q_\theta}{p}\right) +\lambda\right)\deltheta \log q_\theta \right]. \label{eq:dfexp}
\end{equation}
Alternatively, using a proposal distribution $q^\prime$ with full support in $\mathcal{A}$, the expectation \revise{in }\eqref{eq:dfexp} can be reformulated as
\begin{equation}
    \deltheta\df = \mathbb{E}_{q'}\left[\frac{q_\theta}{q'} \left(f'\left(\frac{q_\theta}{p}\right) +\lambda\right)\deltheta \log q_\theta \right]. \label{eq:dfresampl}
\end{equation}

\subsection{Gradient Estimation}
\label{sec:gradEst}
Given a sample $a\sim q_\theta$, it is not possible to directly evaluate $q_\theta(a)$ as it is not available in closed form. Therefore, $q_\theta$ needs to be estimated to compute the gradients of the f-divergence.
Given $N$ sampled actions ${a_i} \sim q_\theta$, $q_\theta$ can be approximated with a \gls*{kde} by
\begin{equation}
    q_\theta(a)\approx \hat{q}_{\theta, \sigma}(a) = \frac{1}{N}\sum_{a_i \sim q_\theta}k_\sigma(a - a_i),
    \label{eq:kde}
\end{equation}
where $k_\sigma$ is a Gaussian kernel with a diagonal bandwidth matrix $\sigma$. The \gls*{kde} enables the estimation of the expectation. Using \eqref{eq:dfexp}, computing the expectation value as the average over the samples yields
\begin{equation}
    \deltheta\df \approx \frac{1}{N}\sum_{a_i\sim q_\theta}\left(f' \left(\frac{\hat{q}_{\theta, \sigma}}{p}\right) + \lambda\right)\deltheta\log \hat{q}_{\theta, \sigma}.
    \label{eq:est_df}
\end{equation}
\revise{The gradient estimator in} \eqref{eq:est_df} did not converge in our experiments. While a systematic investigation of the convergence issue was not completed, we suspect two primary reasons. First, the support $q_\theta$ usually does not cover the whole action space $\mathcal{A}$, which is necessary for the expectation formulation in \eqref{eq:dfexp}. Second, evaluating $q_\theta(a_i)$ based on a \gls*{kde}, which uses ${a_j}$ as supports, has a bias for $j=i$. 

Adding Gaussian noise to the samples gives full support in $\mathcal{A}$ and reduces the bias at the support points of the KDE, which led to convergence in the experiments. The new samples are given by $a^*_j = a_i + \epsilon$ for $mi \leq j < m(i+1)$ and $\epsilon \sim \mathcal{N}(0,\sigma')$,
where $m$ indicates the number of samples drawn for each original sample. This is equivalent to sampling from a \gls*{kde} with $a_i$ as supports and $\sigma^\prime$ as bandwidth. \revise{Using importance sampling in} \eqref{eq:dfresampl}\revise{, the gradient in}~\eqref{eq:est_df}\revise{ after resampling  can be rewritten as follows }
\begin{align}
    & \label{eq:est_dfresampl}
    \deltheta\df \approx \nonumber \\ & \frac{1}{M}\sum_{a^*_j\sim \hat{q}_{\theta, \sigma^\prime}}\frac{\hat{q}_{\theta, \sigma}}{\hat{q}_{\theta, \sigma^\prime}}\left(f' \left(\frac{\hat{q}_{\theta, \sigma}}{p}\right) + \lambda\right)\deltheta\log \hat{q}_{\theta, \sigma},
\end{align}
with $M = mN$. Additionally, \revise{equation}~\eqref{eq:est_dfresampl} requires an estimate of $p$, which in turn requires an estimate of the volume in \eqref{eq:scoredist}
\begin{equation}
    \intA{g(a)} \approx
    \frac{1}{M}\sum_{a^*_j}\frac{g(a^*_j)}{\hat{q}_{\theta, \sigma^\prime}(a^*_j)}.
    \label{eq:mcis}
\end{equation}
\revise{This volume estimation in}~\eqref{eq:mcis} is similar to self-normalized importance sampling (\cite{Murphy2012}) but uses the proposal distribution. The bandwidth $\sigma^\prime$ of the proposal distribution is a hyper-parameter. Setting $\sigma^\prime = c \,\sigma$, experiments show that in most cases $c > 1$ helps convergence. Intuitively, a larger bandwidth enables the exploration of nearby modes in the action space. 
Specific estimators for the different f-divergences can be obtained by substituting $f'$ from Table~\ref{tb:divergences} into \eqref{eq:est_dfresampl}. A summary of the gradient estimators used in this work is given in Table~\ref{tab:grad_est}.

\begin{table}
    \centering
    \caption{Gradient estimators of various losses and choice of Lagrangian multiplier $\lambda$.}
    \setlength{\extrarowheight}{5pt}
    \begin{tabular}{c|l|c}
        Loss & Actor Gradient Estimator & $\lambda$\\ \hline
        
        \acrshort*{js} & $\frac{1}{2M}\sum_{a_j^*} \frac{\hat{q}_{\theta, \sigma}}{\hat{q}_{\theta, \sigma^\prime}} \log\left(\frac{2\hat{q}_{\theta, \sigma}}{p + \hat{q}_{\theta, \sigma}}\right) \deltheta\log \hat{q}_{\theta, \sigma}$ & 0\\
        
        \acrshort*{fkl} & -$\frac{1}{M}\sum_{a_j^*} \frac{p}{\hat{q}_{\theta, \sigma^\prime}} \deltheta \log \hat{q}_{\theta, \sigma}$ & 0\\
        
        \acrshort*{rkl} & $\frac{1}{M}\sum_{a_j^*} \frac{\hat{q}_{\theta, \sigma}}{\hat{q}_{\theta, \sigma^\prime}} \log\left(\frac{\hat{q}_{\theta, \sigma}}{p}\right) \deltheta\log \hat{q}_{\theta, \sigma}$ & -1\\
        
        \acrshort*{gan} & $\frac{1}{N}\sum_{a_i}\frac{\partial}{\partial a}\log(1 - \xi_\phi)\deltheta a_i$ & - \\
        
        \acrshort*{me} & $\frac{1}{N}\sum_{a_i}\deltheta\log\hat{q}_{\theta, \sigma} - \frac{\partial}{\partial a}\log\xi_\phi\deltheta a_i$ & -\\
        
    \end{tabular}
    \label{tab:grad_est}
\end{table}

\subsection{Training Process}
Algorithm \ref{alg:alg} shows a training loop when training a feasibility policy directly on the feasibility model using a Jensen-Shannon (JS) loss. The training iterates as follows: A batch of random states is sampled, and the actor generates $N$ actions $a_i$ per state. For each action $a_i$, $m$ values are sampled from a normal distribution $\mathcal{N}(0, \sigma^\prime)$ and added to the action values to create the $M$ action samples $a_j^*$. Using the actions $a_i$ as support of the KDE in \eqref{eq:kde}, the densities $\hat{q}_{\theta,\sigma}(a_j^*)$ and $\hat{q}_{\theta,\sigma^\prime}(a_j^*)$ are computed. Then the feasibility model $g$ is evaluated on all samples $a_j^*$ and the estimate of $p(a_j^*)$ is computed using \eqref{eq:scoredist} and importance sampling in \eqref{eq:mcis}. Finally, the gradient of $\theta$ can be computed according to \eqref{eq:est_dfresampl}. For a better understanding of the gradient, the trace of the gradient is highlighted in red throughout the algorithm.

\begin{algorithm}
\caption{Jensen-Shannon training loop} 
\label{alg:alg}
Initialize $\theta$\\
\For{$1$ \KwTo Training Steps }{
    \For{$k=1$ \KwTo K}{
    $s_k \leftarrow$ Sample from $\mathcal{S}$\\
    $z_i \leftarrow $ Sample uniformly in $\mathcal{Z}, \quad\forall i\in[1,N]$ \\
    $\textcolor{red}{a_i} \leftarrow \pi_{\textcolor{red}{\theta}}(s_k, z_i), \quad\forall i\in[1,N]$\\
    $\epsilon_j\sim \mathcal{N}(0, \sigma^\prime), \quad \forall j\in[1, M]$\\
    $a^*_j \leftarrow \operatorname{stop\_gradient}(a_{\lceil j/m \rceil}) + \epsilon_j, \quad \forall j\in[1, M]$\tcp*[f]{Resample from KDE}\\
    $\textcolor{red}{\hat{q}_j} \leftarrow \frac{1}{N}\sum_{i=1}^N k_\sigma(a^*_j - \textcolor{red}{a_i}), \quad \forall j\in[1, M]$\tcp*[f]{Evaluate KDE on samples}\\
    $\hat{q}^\prime_j \leftarrow \frac{1}{N}\sum_{i=1}^N k_{\sigma^\prime}(a^*_j - a_i), \quad \forall j\in[1, M]$\tcp*[f]{Evaluate proposal pdf}\\
    $\hat{r}_j \leftarrow g(s_k, a_j^*),\quad\forall j\in [1,M]$\tcp*[f]{Evaluate feasibility model on samples}\\
    $\hat{V}\leftarrow \frac{1}{M} \sum_{j=1}^M \frac{\hat{r}_j}{\hat{q}^\prime_j}$\tcp*[f]{MC integration with importance sampling}\\
    $\hat{p}_j\leftarrow \frac{\hat{r}_j}{\hat{V}},\quad\forall j\in [1,M]$\\
    $g_k \leftarrow \frac{1}{2M}\sum_{j=1}^{M}\frac{\hat{q}_j}{\hat{q}^\prime_j} \log\left(\frac{2\hat{q}_j}{\hat{q}_j + \hat{p}_j}\right) \nabla_\theta \log(\textcolor{red}{\hat{q}_j}) $ \tcp*[f]{\textcolor{red}{gradient trace}}
    }
    $\theta \leftarrow \theta - \alpha_\theta \frac{1}{K}\sum_{k=1}^K g_k$
}
\end{algorithm}
Intuitively, the gradient in \eqref{eq:est_dfresampl} attracts support actions $a_i$ towards sample actions $a_j^*$ where $p(a_j^*) > \hat{q}_{\theta,\sigma}(a_j^*)$ and repulses support actions from samples where $p(a_j^*) < \hat{q}_{\theta,\sigma}(a_j^*)$. The different f-divergences place different weights on attraction and repulsion. FKL only attracts support actions towards samples with high $p$, while RKL repulses strongly from samples with $p=0$, and JS attracts and repulses with lower magnitude.

\subsection{Actor-Critic}\label{sec:method:ac}
Algorithm \ref{alg:alg} assumes that the training can be performed directly on the feasibility model. However, multiple actions must be evaluated for the same state to train the actor. This is possible if $g$ is available in closed form or effectively simulated. In some scenarios, $g$ can be a real experiment that does not allow reproducibility of states. To mitigate this problem, an auxiliary neural network 
$
    \xi_\phi : \mathcal{S} \times \mathcal{A} \to \mathbb{R}
$
with parameters $\phi$ can be trained to imitate the environment $g$. The policy can then be trained to match the distribution of feasible actions according to this auxiliary neural network. We refer to $\pi_\theta$ and $\xi_\phi$ as actor and critic, respectively. 

The actor and critic can be trained simultaneously. The critic is trained on data from a replay memory collected through interactions between the actor and the environment, with each training batch containing half feasible actions and half infeasible actions to stabilize training. To further improve the training efficiency of the critic, when the actor interacts with the environment, it suggests multiple actions, which the critic evaluates. The action with the highest uncertainty, i.e., the action with $\xi\approx 0.5$ is selected as it contains the most information for the critic. We call this process \textit{maximum uncertainty sampling}. During evaluation, to improve the precision of the actor, the critic can be evaluated on proposed actions, and actions with low values can be rejected. This action optimization can increase precision but may reduce recall or the ability to find all the disconnected sets of feasible actions.
\section{Illustrative Example}
\label{sec:illustrative}
\begin{figure*}
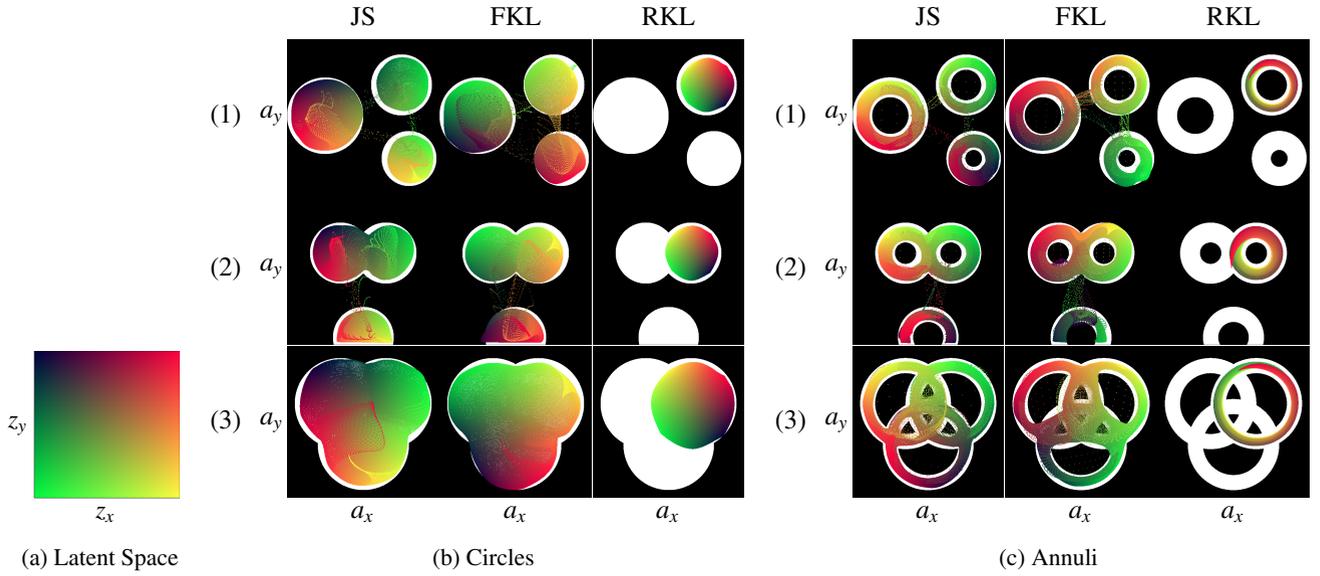

    \centering
    \begin{subfigure}{0.15\textwidth}
            \resizebox{1.0\textwidth}{!}{%
    \input{images/circles/latent}
  }
  \caption{Latent Space}\label{fig:circles:latent}
    \end{subfigure}%
    \begin{subfigure}{0.42\textwidth}
            \resizebox{1.0\textwidth}{!}{%
    \input{images/circles/circles}
  }
  \caption{Circles}
    \end{subfigure}%
    \begin{subfigure}{0.42\textwidth}
            \resizebox{1.0\textwidth}{!}{%
    \input{images/circles/donuts}
  }
  \caption{Annuli}
    \end{subfigure}

    \caption{Illustrative example showing two feasibility models, which specify feasible regions as the union of three random circles (b) or annuli (c). Three states (1)-(3) are shown for each example, solved with the JS, FKL, and RKL divergence, with feasible and infeasible action space in white and black, respectively. The colored points are actions generated by the feasibility policy when using the corresponding latent space value $(z_x, z_y) \in \mathcal{Z}$ in (a).}
    \label{fig:example}
\end{figure*}
This section provides illustrative examples to elucidate the feasibility policy and demonstrates the potential for direct training on a parallelizable feasibility model across multiple actions for a given state. Hyperparameters, their ranges, \revise{and training and inference times} are summarized in Table \ref{tab:params}.

\subsection{Problem}
Consider three circles with given radii and center points as the state $s$. The feasibility model $g$ deems any point $\mathbf{a}$ a feasible action if it falls within at least one circle and lies inside a unit square, described as follows: $s = {(\mathbf{c_k}, r_k)}_{k\in{1,2,3}}$ where $(\mathbf{c_k}, r_k)$ are the center points and radii of the circles, and $\mathbf{a}\in\mathbb{R}^2$ represents a coordinate. The feasibility model is thus expressed by
\begin{equation}
g(s,\mathbf{a}) = (\mathbf{0}\leq \mathbf{a} \leq \mathbf{1})\land \bigvee_{k=1}^3 (|\mathbf{a} - \mathbf{c_k}| < r_k).
\end{equation}
In the extended example, each circle includes an inner radius, forming annular regions.

\subsection{Results}

Figure \ref{fig:example} illustrates the outcomes of applying three distinct divergences, JS, FKL, and RKL, to the circle and annulus scenarios, depicted in subfigures (b) and (c), respectively. Actions are generated from a grid of $256^2$ latent values shown in subfigure (a), where each color corresponds to a specific latent value. Three states, marked as (1), (2), and (3), represent various configurations: disconnected shapes, partially connected shapes, and fully connected shapes. The figure visually underscores the different outcomes using the divergences: the RKL approach tends to focus on singular modes, even failing to span overlapping regions, as seen in the third row of both (b) and (c). On the contrary, both FKL and JS exhibit a more expansive coverage, approaching the borders of the feasible space, indicated by the white regions, with the JS divergence showing a reduced density within the infeasible space, represented by the black regions, as compared to FKL. This phenomenon is particularly evident in the first and second states for the circle and annulus examples, which can be attributed to the repulsive gradient present in JS divergence that is absent in the FKL divergence.

These visualizations show that a feasibility policy can be trained to navigate complex distributions beyond the Gaussian reparameterization commonly found in the literature. They further elucidate the importance of enabling the FKL and JS divergences to address disconnected feasible sets effectively. Ultimately, these examples offer an intuitive comprehension of the aim: for the feasibility policy to generate all feasible actions by learning to map the latent space into diverse shapes conditioned on the state.
\section{Feasible Trajectory Segments Example}
\label{sec:splines}

\begin{figure*}
    \centering
    \begin{subfigure}{0.24\textwidth}
        \includegraphics[width=\textwidth]{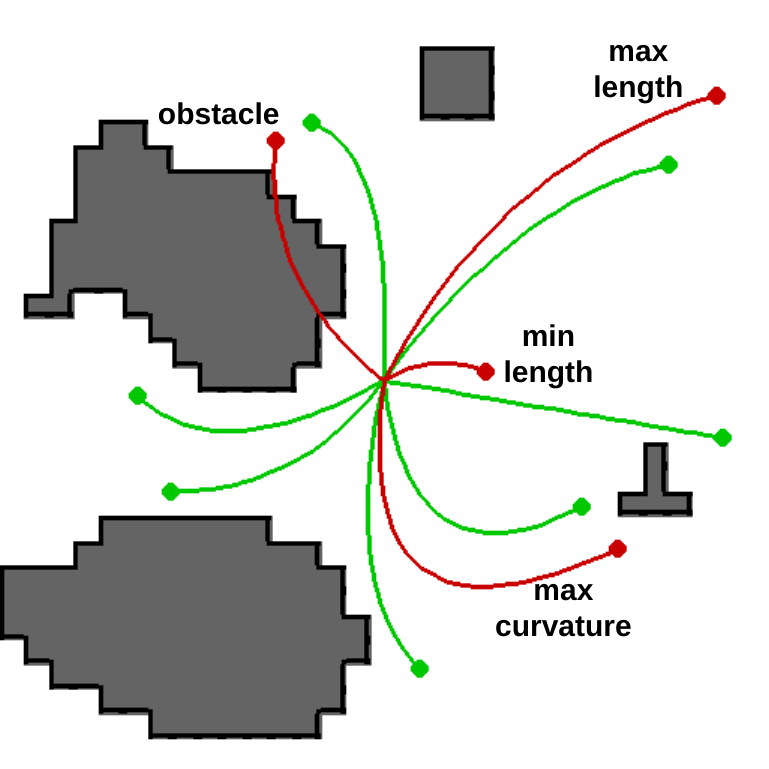}
        \caption{Example actions on map 1}
        \label{fig:constraints}
    \end{subfigure}\hspace{1pt}
    \begin{subfigure}{0.24\textwidth}
        \includegraphics[width=\textwidth]{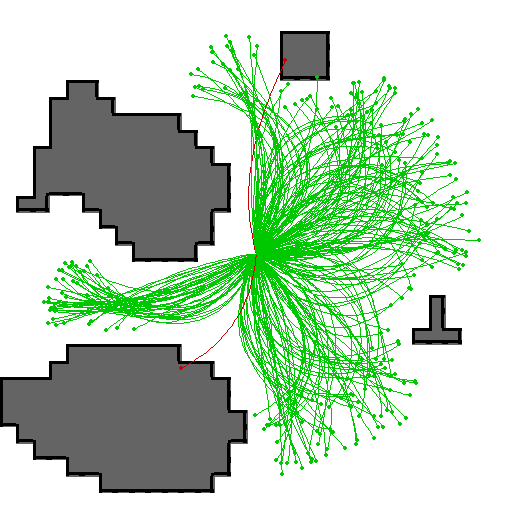}
        \caption{Map 1}
        \label{fig:map1}
    \end{subfigure}\hspace{1pt}
    \begin{subfigure}{0.24\textwidth}
        \includegraphics[width=\textwidth]{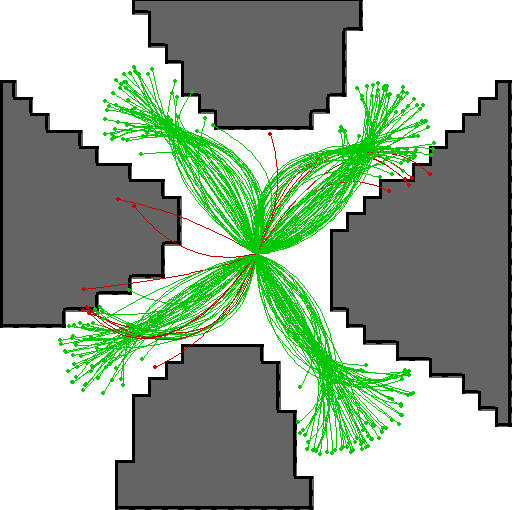}
        \caption{Map 2}
        \label{fig:map2}
    \end{subfigure}\hspace{1pt}
    \begin{subfigure}{0.24\textwidth}
        \includegraphics[width=\textwidth]{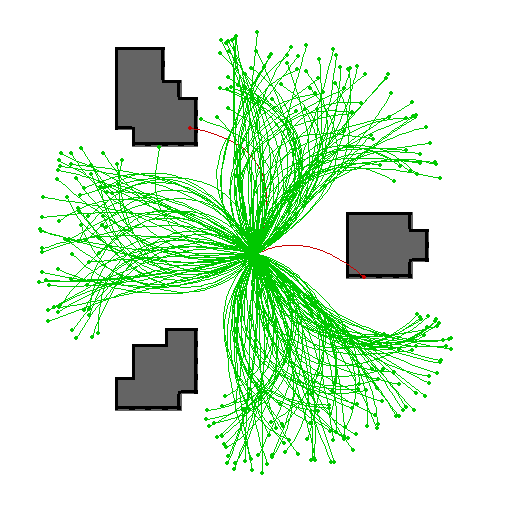}
        \caption{Map 3}
        \label{fig:map3}
    \end{subfigure}
    \caption{\revise{Quadratic spline action space application showing three different maps: a randomly generated map in (a) and (b) and two handcrafted maps in (c) and (d). In (a), example splines are shown with green indicating a feasible spline and red indicating an infeasible one. An example for each constraint violation is given. In (b)-(d), the agent generates 256 actions that are displayed with the color depending on the feasibility of each proposed action.}}
    \label{fig:splines}
\end{figure*}
\revise{
When solving problems in robotic path planning with reinforcement learning, a standard action space is the direction and velocity target of the robot. However, in tasks that span a long time horizon, it can be beneficial to reduce the number of actions by bundling multiple actions in parametric trajectory segments, often splines, to be followed. Another benefit of generating splines is that these can be checked for collisions with obstacles and other system constraints, such as maximum curvature. This application example shows how learning all feasible actions could be used in this context. 
}
\subsection{Problem}
\revise{
Consider a stationary agent  at the center of an environment with known obstacles. In this example, the objective is to find all quadratic splines that fulfill the following conditions}
\begin{enumerate}
    \item \revise{does not intersect with any obstacle;}
    \item \revise{longer than a minimum length;}
    \item \revise{shorter than a maximum length;}
    \item \revise{its maximal curvature is less than a threshold.}
\end{enumerate}
\revise{Figure}~\ref{fig:constraints} \revise{shows an example scenario with randomly generated obstacles (gray) and example splines. For each constraint, the figure shows an example that violates it, additionally providing examples of feasible splines. The splines are parameterized through the endpoint and an intermediary point that bends the spline, yielding a 4D action space. The feasibility model checks for any constraint violation numerically along the spline. The agent observes the obstacles as a black and white image with size $31\times 31$. It is trained with the JS loss on randomly generated obstacle maps and evaluated on maps not seen during training. The parameters for training, and training and inference times are given in Table}~\ref{tab:params}.

\subsection{Results}
\revise{Figure}~\ref{fig:splines} \revise{shows three example obstacle maps and action samples from the agent. In Figure}~\ref{fig:map1}\revise{, the agent provides 256 splines for the randomly generated map, among which 254 are feasible. On the right side of the map, with only two smaller obstacles, the agent produces a wide range of splines that avoid the two obstacles, with a larger margin toward the bottom obstacle. The left side of the map, with larger obstacles and only a smaller gap for feasible paths, shows that the agent also produced a group of splines. Given the minimum length constraint on the splines, the splines going to the left are disconnected from the splines on the right, considering the parameter space. The two infeasible splines generated by the agent are likely to be on the transition boundary between these disconnected sets of feasible splines. }

\revise{Map 2 in Figure}~\ref{fig:map2} \revise{shows a situation that contains four disconnected sets of feasible splines, one in each diagonal direction. The agent produces feasible splines in each direction, though generating more infeasible splines. This is likely due to the difficulty of generating four relatively small disconnected sets separated by large volumes of infeasible action space. Map 3 in Figure}~\ref{fig:map3} \revise{shows a simpler problem with three small obstacles resulting in three disconnected sets of feasible splines. In this example, the agent again generated 254 feasible splines in all three sets with only two splines when transitioning between sets. Overall, the agent can generate splines in all disconnected sets, largely avoiding generating infeasible splines. }

\revise{This example shows how action mapping could be applied to motion or path planning problems when they are solved with reinforcement learning. It can be clearly seen that the feasibility policy learned to generate splines representative of all feasible options with only a few infeasible splines. Therefore, an objective policy should be greatly aided if it only needs to choose among the splines that the feasibility policy can generate. In our future work, we plan to investigate action mapping using splines as action space in a reinforcement learning-based path planning problem.}

\section{Robotic Grasping Setup}
\label{sec:setup}

Besides the illustrative examples, the proposed method was tested in a simplified robotic grasping simulation, where we compare different f-divergences with other approaches and investigate how the proposed approach reacts to distortions in the observation.

\subsection{Grasping Simulation}
 Our grasping simulator generates four shapes (H, 8, Spoon, T) for training and a Box shape for testing. The shape position, orientation, color, and geometry parameters are randomly sampled, producing various observations. The observation space is a $128\times 128$ pixel RGB image. We assume a vertical configuration of a parallel gripper with three degrees of freedom $x$, $y$, and $\alpha$ and assume that the object is an extrusion of the 2D shape in the observation. The action space is constrained to the center $78\times 78$ pixel region to avoid undefined behavior at the border of the RGB image. The angle of the grasp is in $[0, \pi)$ as the gripper is symmetrical; thus, a complete revolution is unnecessary.

\begin{figure*}
    \centering
    \begin{tikzpicture}
        \coordinate (center) at (0,0);
        \node[above = of center, yshift=-2.7em]  {\includegraphics[width=0.98\textwidth]{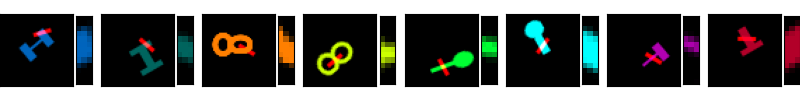}};
        \node[] at (-7.8, 0) {\small$(a)$};
        \node[] at (-5.57, 0) {\small$(b)$};
        \node[] at (-3.34, 0) {\small$(c)$};
        \node[] at (-1.11, 0) {\small$(d)$};
        \node[] at (1.11, 0) {\small$(e)$};
        \node[] at (3.34, 0) {\small$(f)$};
        \node[] at (5.57, 0) {\small$(g)$};
        \node[] at (7.8, 0) {\small$(h)$};
    \end{tikzpicture}%
    \caption{Feasible gripper positions (red) for different variations of the shapes (\textit{H-shape} (a+b), \textit{8-shape} (c+d), \textit{Spoon} (e+f), and \textit{T-shape} (g+h)) used in training, with a detailed view of the area between the gripper to the right of each figure.}
    \label{fig:gripper_positions}
\end{figure*}

The success of a grasp is only determined by the relative position and alignment of the gripper to the outline of the object, as seen from a camera positioned above the experiment. Given the alignment of the gripper, i.e., $x$, $y$, and $\alpha$ and a simulated picture of the object from a fixed camera, we developed an algorithm that provides a success/failure outcome in a deterministic and reproducible manner. Given the maximum aperture of the parallel gripper $l$ and the width of the gripper claws $w$, the simulation analyzes the cropped image content of dimensions $l\times w$ between the gripper claws before the claws close on the object. The simulation checks if the object is sufficiently present, equidistant from the claws, and aligned within parameterized margins. Figure~\ref{fig:gripper_positions} shows successful grasping poses and the respective gripper content for the objects that are trained on.

\begin{figure}
    \centering
    \includegraphics[width=\columnwidth]{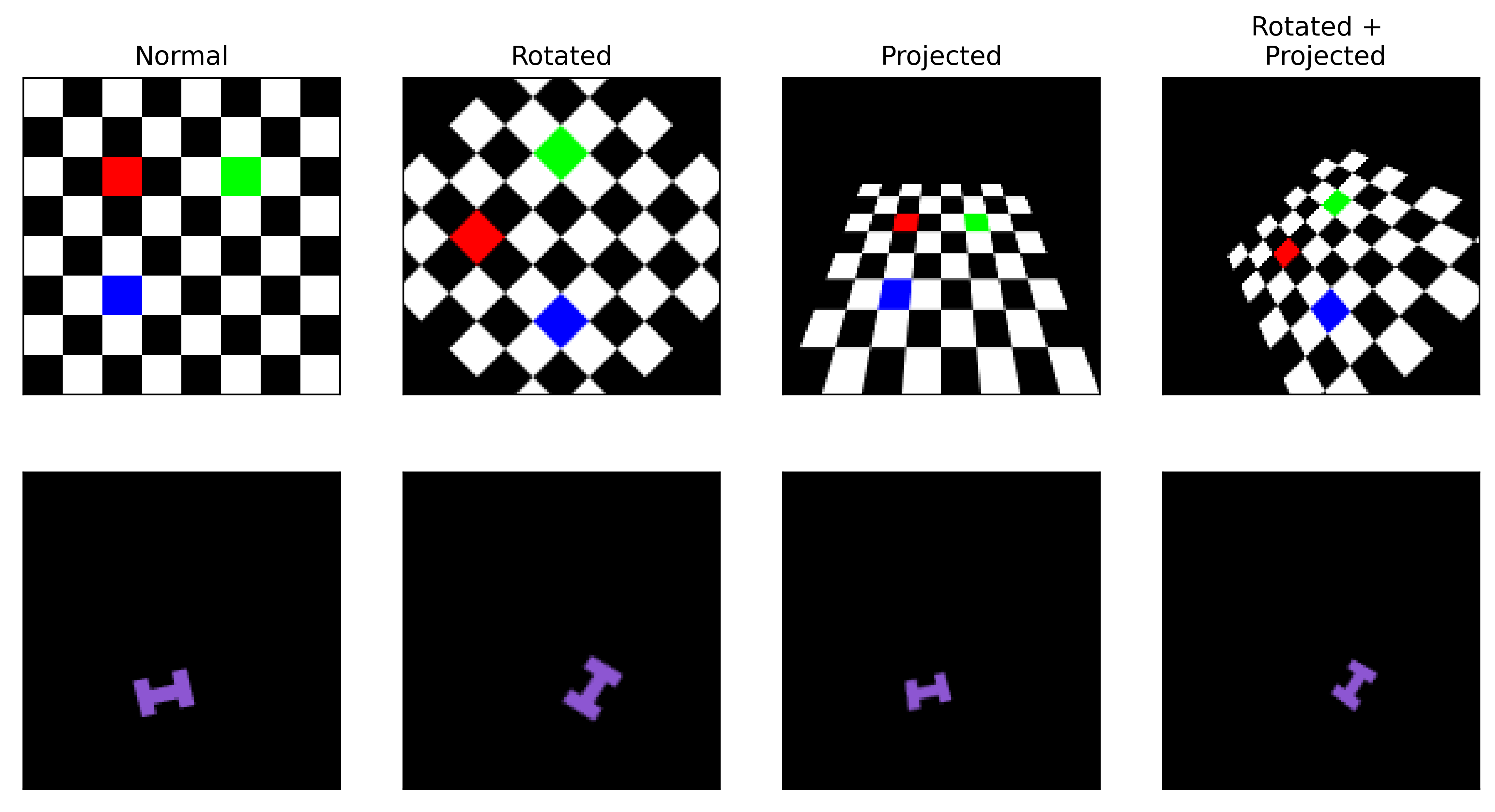}
    \caption{Different distortions are applied, showing a colored chess board for illustration and an example shape under all distortions.}
    \label{fig:distort}
\end{figure}
In the primary experiment, we test the algorithm under aligned observation and action spaces. In a second study, we investigate if distortions of the observation affect the performance. The distortions investigated are a rotation, projection, and rotation + projection as shown in Figure~\ref{fig:distort}. These distortions correspond to different camera perspectives. The applied distortion is only on the observation and does not change the mechanics of the experiment.

\subsection{Neural Network Design}

\begin{figure}
    \centering
    \resizebox{\columnwidth}{!}{
    \begin{tikzpicture}
    \input{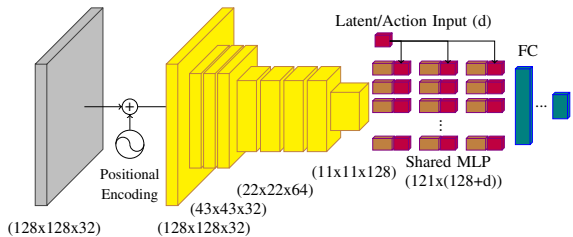}
    \end{tikzpicture}}
    \caption{Before processing, the image is embedded (in gray) and augmented with positional encoding, resulting in 32 total channels. After positional encoding, a convolutional layer with stride 3, followed by 7 residual blocks (in yellow) with a bottleneck, preprocesses the state. The output is processed by 3 layers of "pixel-wise" shared MLPs (in brown), with the features being concatenated with a latent input (in purple) of length $d$. The latent input is a random sample from $\mathcal{Z}$ for the actor and the action to be evaluated for the critic. Four (for the actor) or three (for the critic) fully connected layers (in blue) output the action and the feasibility estimate, respectively.}
    \label{fig:neuralnetwork}
\end{figure}

The neural network that was used for the actor and critic in the robotic experiment is illustrated in Figure~\ref{fig:neuralnetwork}. The neural network design was guided by simplicity and inspired by \glspl*{gan}. Features that rely on domain-specific knowledge are avoided to evaluate better the learning method presented in the paper. The actor and critic share the residual feature extraction network (\cite{He2016}). The hyperparameters for training \revise{and training and inference times} are summarized in Table \ref{tab:params}.

As a peculiarity of the network and the loss, the actor's inferred action has four components, $[x,y,r\sin{\alpha}, r\cos{\alpha}]$, with $r\in [0,\sqrt{2}]$. The angle can be extracted trivially with the $\operatorname{arctan}$ of the ratio of the third and fourth action components. As the scale factor $r$ does not change the angle, the critic receives the normalized action $[x,y,\sin{\alpha}, \cos{\alpha}]$ as input. To avoid the actor from reaching the singularity at $r=0$ and the distribution $q$ being spread along the radius, $g(s,a)$ and $\xi(s,a)$ are scaled with an unnormalized Gaussian on the radius, centered at $0.5$ with the standard deviation of $\sigma_{sc}$.

\begin{table}
    \centering
    \caption{\revise{List of parameters for all experiments.}}
    \footnotesize
    \setlength{\tabcolsep}{1pt}
    \begin{tabular}{c|c|c|c|l}
     & Sec. \ref{sec:illustrative} & Sec.~\ref{sec:splines}& Sec. \ref{sec:setup} & Description \\ \hline
        N & 128 & 256 & 128 & Support size \\
        M & 256 & 256 & 256 & Resampling size \\
        $\sigma_{xy}$ & 0.01 & 0.1 & 0.025 & \gls*{kde} bw. $x,y$\\
        $\sigma_{sc}$ & - & - & 0.4 (RKL: 0.1) & \gls*{kde} bw. $\sin\alpha,\cos\alpha$\\ 
        $c$ & 2.0 & 2.0 & 3.0 & Sampling bw. $\sigma^\prime = c\sigma$\\ \hline
        U & - & - & 64 & Max Uncertainty Proposals \\
        $|\mathcal{M}|$ & - & - & 320,000 & Replay memory size \\ \hline
        K & 16 & 16 & 16 & Actor batch size \\
        L & - & - & 32 & Critic batch size \\
        $lr_a$ & $5 * 10^{-5}$ & $5 * 10^{-5}$ & $5 * 10^{-5}$ & Learning rate actor \\
        $lr_c$ & - & - & $5 * 10^{-5}$ & Learning rate critic \\ \hline
        $\mathbf{c}$ & $[0.0, 1.0]^2$ & - & - & Center point range \\
        $r$ & $[0.1, 0.3]$ & - & - & Radius range circles \\
        $r_{o}$ & $[0.2, 0.3]$ & - & - & Outer radius range annuli \\
        $r_{i}$ & $[0.3, 0.7] * r_{o}$ & - & - & Inner radius range annuli \\ \hline
        $l_\text{min}$ & - & 0.5 & - & Minimum spline length \\
        $l_\text{max}$ & - & 1.0 & - & Maximum spline length \\
        $c_\text{max}$ & - & 8.0 & - & Maximum curvature \\ \hline
        $T_t$ & 8 h & 4 h & 48 h & Training time \\
        $I_1$ & 2.2 ms & 1.5 ms & 4.0 ms & Inference time 1 action \\
        $I_{256}$ & 2.2 ms & 1.5 ms & 8.3 ms & Inference time 256 actions \\ 
        $I_{4096}$ & 9.0 ms & 4.0 ms & 92.0 ms & Inference time 4096 actions \\ \hline
        \multicolumn{5}{p{240pt}}{ \revise{The training and inference times were measured on an NVIDIA A100 GPU. Inference times for multiple actions measure generating multiple actions for one problem. RKL is sensitive to large KDE bandwidths and benefits from a smaller bandwidth for $\sin\alpha,\cos\alpha$.}}
    \end{tabular}
    \label{tab:params}
\end{table}

\subsection{Comparison}

In the primary experiment, we are comparing different f-divergences with each other and with two other approaches. The first is a \gls*{me} RL algorithm similar to \gls*{sac} in \cite{haarnoja2018soft}, which trains the actor to minimize
\begin{equation}
    \min_\theta \mathbb{E}_{s \sim \mathcal{M}, z\sim \mathcal{Z}}\left[ \log q_\theta(\pi_\theta(s,z)|s) - \xi_\phi(s, \pi_\theta(s,z))  \right],
\end{equation}
with $\mathcal{M}$ being the replay memory. The critic is trained as described in Section~\ref{sec:method:ac}. Instead of using the reparameterization trick with a known distribution to estimate the entropy, we use the \gls*{kde}. The other approach is an implementation of a conditional \gls*{gan} (\cite{mirza2014}) with a growing dataset. The min-max optimization problem is given through
\begin{equation}
    \min_\theta \max_\phi \mathbb{E}_{\substack{s,a \sim \mathcal{M}_p\\ z\sim \mathcal{Z}}}\left[ \log( \xi_\phi(s,a)) - \log (1 -\xi_\phi(s,\pi_\theta(s, z))) \right],
\end{equation}
with a positive replay memory $\mathcal{M}_p$ only containing feasible actions. An asterisk is added (e.g., JS*) when using action optimization, rejecting $10\%$ of the proposed actions with the lowest critic value. 

In the secondary evaluation, we compare with a common approach in the literature (\cite{zeng2020tossingbot}) that uses spatial equivariance. The domain-specific approach utilizes fully convolutional networks to output a probability of success for each action of a \textit{discretized} action space. As in \cite{zeng2020tossingbot}, the observation is fed into the neural network multiple times with different rotations. The neural network then only needs to output a one-channel image containing the probability of success of each discretized $x,y$ action for the given rotation of the image. This approach thus uses translation equivariance by using a \gls*{cnn} and rotation equivariance. In the experiments, we denote it as the heat-map approach (H). 

The approach is implemented using fully convolutional networks with an hourglass structure, adopting the beginning of the Resnet in Figure~\ref{fig:neuralnetwork} and adding the same structure in reverse order with nearest-neighbor upsampling. The approach predicts grasping success for 78x78 pixels with 16 rotation angles, trained on a cross-entropy loss on the grasping outcome sampled from the replay buffer. The replay buffer is also filled with imitation learning examples, and maximum uncertainty sampling is applied. For evaluation, the success estimate of each discretized action is used as its probability to be sampled. To increase accuracy, an inverted temperature factor increases the difference between higher and lower score actions. Specifically, the actions are sampled according to
\begin{equation}
    q(a|s) = \frac{\operatorname{exp}(\beta\log\xi(s,a))}{\sum_{\forall a \in \mathcal{A}_d}\operatorname{exp}(\beta\log\xi(s,a))},
\end{equation}
with $\xi$ being the fully convolutional network with $s$ as input and as output shape the discretized action space $\mathcal{A}_d$. The inverted temperature was set to $\beta=100$ for $H$ and $\beta=1000$ for $H^*$.
\newpage
\section{Robotic Grasping Results}
\label{sec:results}

\subsection{Top-Down Observation}

For each configuration, three agents were trained for 1 million interaction steps with the environment, taking approximately 48 hours per agent on a single NVIDIA 40GB A100 GPU. At the start of the training, 80k examples, including positives and negatives, for randomly generated shapes were added to the replay memory to bootstrap the critic and discriminator learning. The training architecture is implemented in TensorFlow\cite{tensorflow2015-whitepaper} with the parameters in Table~\ref{tab:params}.

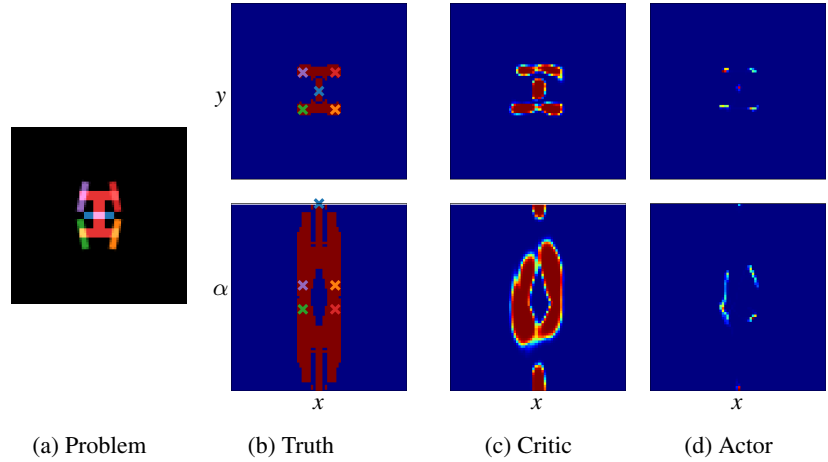
\begin{figure*}
    \centering
    \def\x{0.13}
    \def\xg{0.13}
    \begin{subfigure}[b]{\x\textwidth}
    \begin{tikzpicture}
        \coordinate (center) at (0,0);
        \node[above = of center.270, yshift=0.2cm]  {\includegraphics[width=\textwidth]{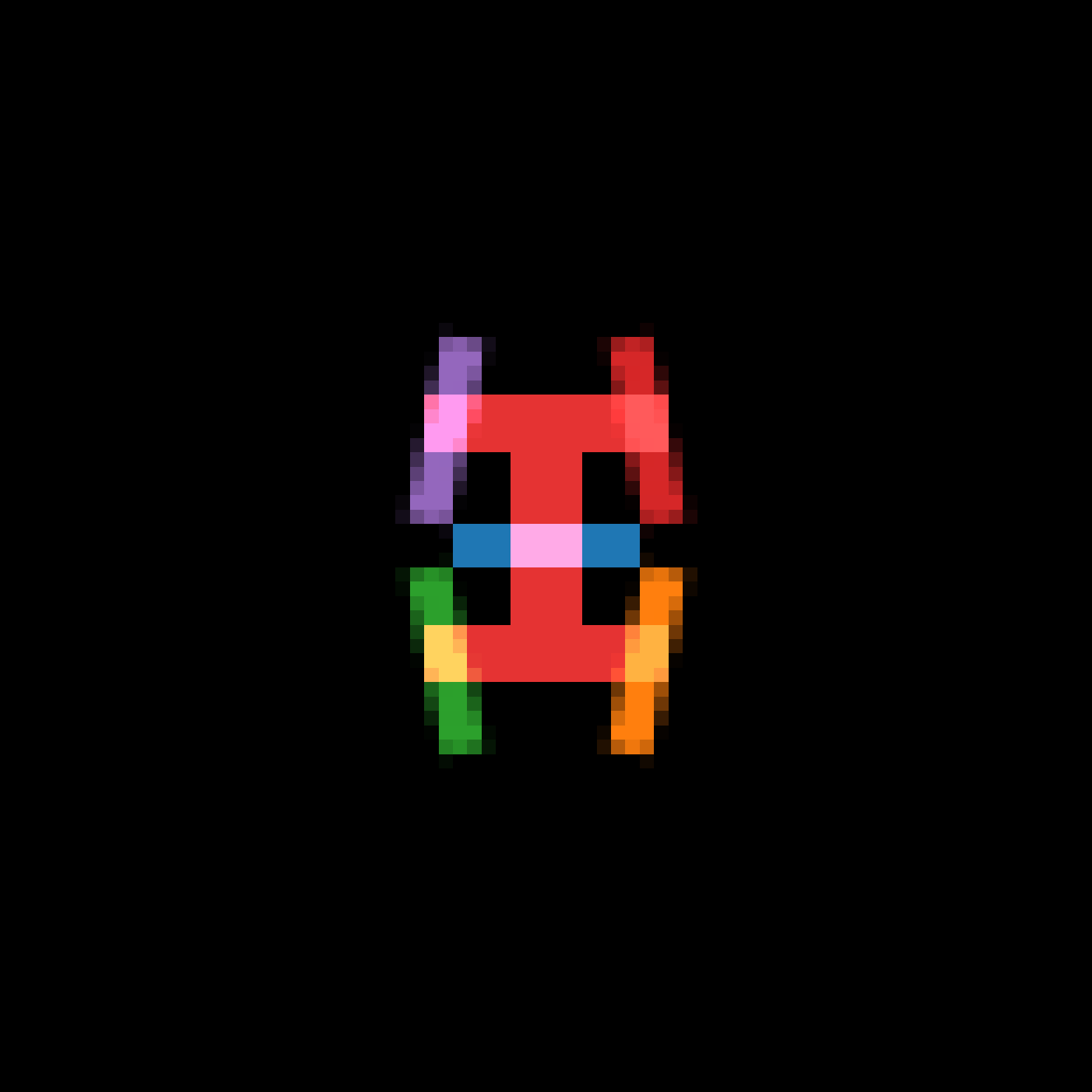}};
        \node[] at (center) {\phantom{\small$x$}};
    \end{tikzpicture}
    \caption{Problem}
    \label{fig:problem:a}
    \end{subfigure}
    \hspace{5pt}
    \begin{subfigure}[b]{\xg\textwidth}
    \begin{tikzpicture}
        \coordinate (center) at (0,0);
        \node[above = of center.270, yshift=-2.7em]  {\includegraphics[width=\textwidth]{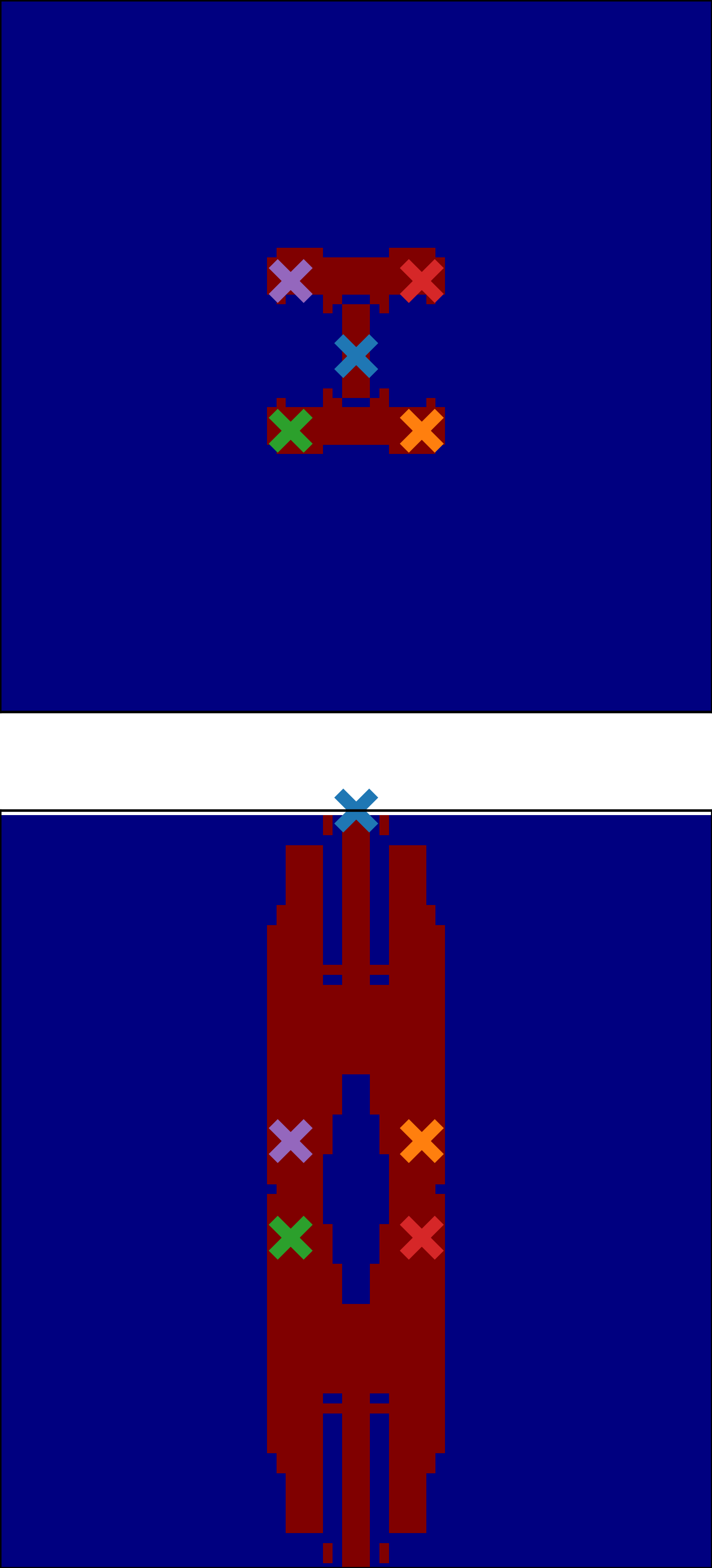}};
        \node[] at (center) {\small$x$};
        \node[] at (-1.3, 1.5) {\small$\alpha$};
        \node[] at (-1.3, 4) {\small$y$};
    \end{tikzpicture}
    \caption{Truth}
    \label{fig:problem:b}
    \end{subfigure}\hspace{21pt}
    \begin{subfigure}[b]{\x\textwidth}
        \begin{tikzpicture}
        \coordinate (center) at (0,0);
        \node[above = of center.270, yshift=-2.7em]  {\includegraphics[width=\textwidth]{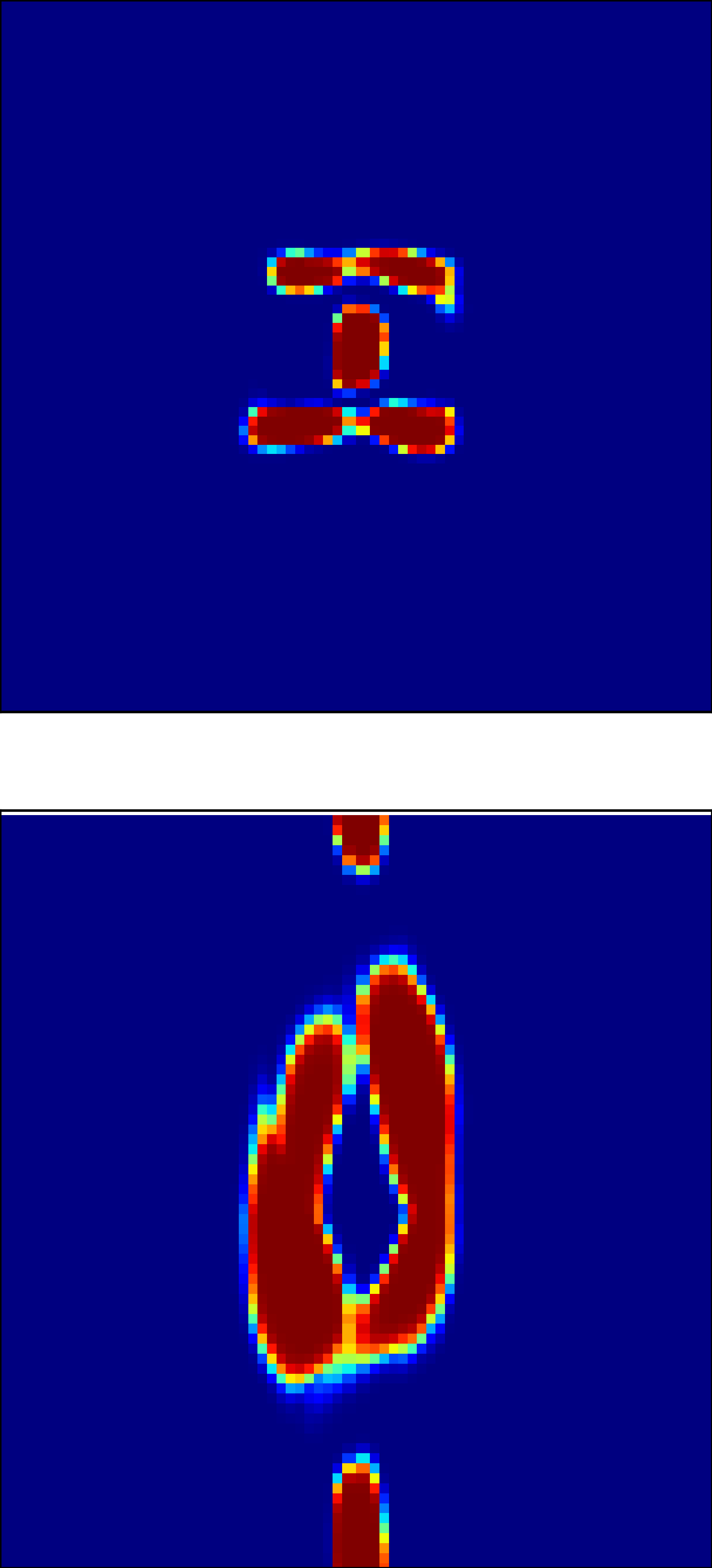}};
        \node[] at (0, 0) {\small$x$};
        \end{tikzpicture}
    \caption{Critic}
    \label{fig:problem:c}
    \end{subfigure}\hspace{7pt}
    \begin{subfigure}[b]{\x\textwidth}
    \begin{tikzpicture}
        \coordinate (center) at (0,0);
        \node[above = of center.270, yshift=-2.7em]  {\includegraphics[width=\textwidth]{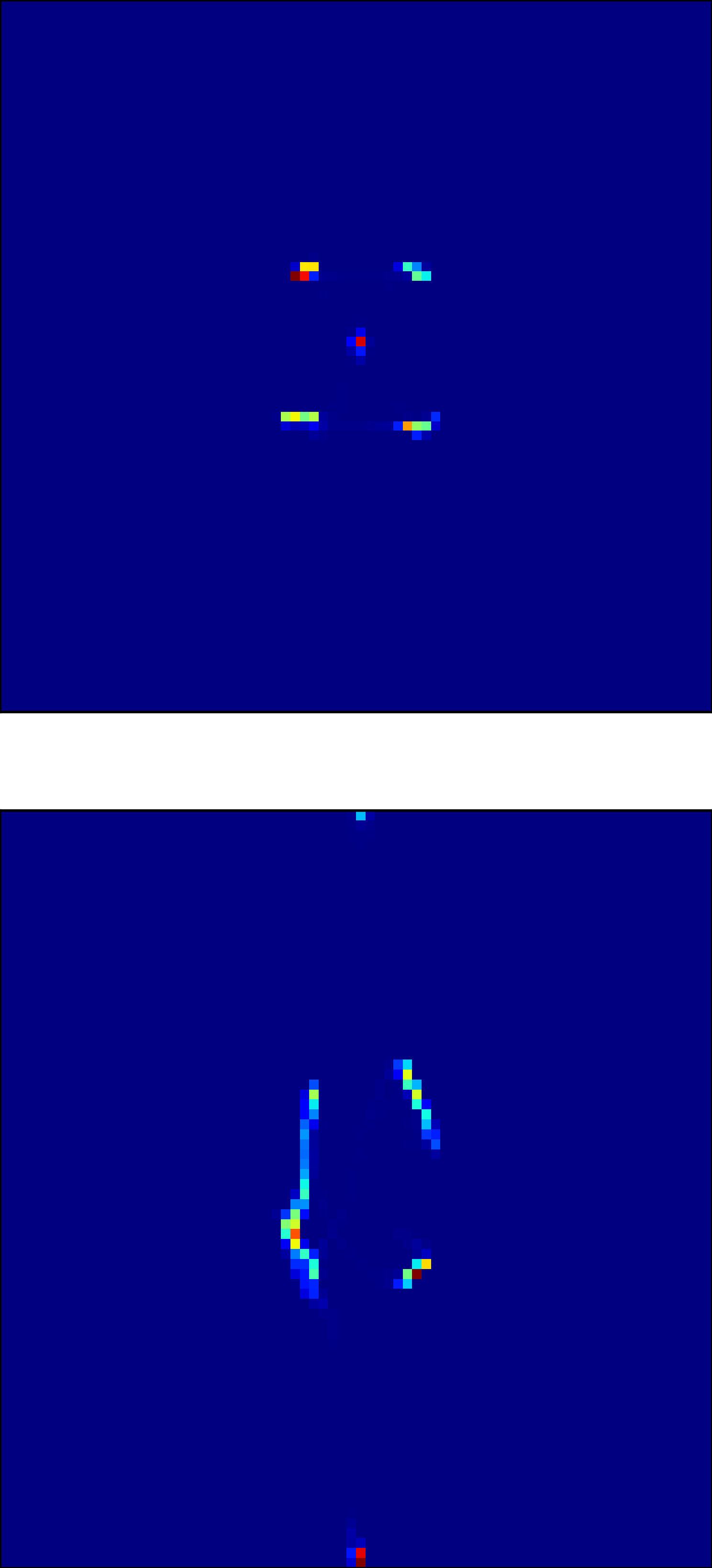}};
        \node[] at (0, 0) {\small$x$};
    \end{tikzpicture}
    \caption{Actor}
    \label{fig:problem:d}
    \end{subfigure}
    \caption{Critic classification and actor distribution trained with JS compared with the ground truth. Five example grasps are shown in the problem and their associated locations in the ground truth. The figures show projections onto the x-y plane (top row) and the x-$\alpha$ plane (bottom row).}
    \vspace{-10pt}
    \label{fig:problem}
\end{figure*}

Figure~\ref{fig:problem} shows the problem, the ground truth feasible picking positions, the critic estimate, and a heat-map of the actor's proposed actions. All figures are projections taking the maximum over the dimension that is not shown. In the problem visualization in Figure~\ref{fig:problem:a}, five feasible picks are shown in different colors, which correspond to the markers in Figure~\ref{fig:problem:b}. These markers highlight the complex multimodality of the problem. While it appears that, e.g., red and purple are in the same mode in the x-y projection, it is visible in the x-$\alpha$ projection that they are not directly connected. Figure~\ref{fig:problem:c} shows that the critic has an approximate understanding of the feasible regions of the action space, showing five modes clearly in the x-y projection. The actor distribution in Figure~\ref{fig:problem:d} also shows all five modes, while the output is significantly sharper in the centers of the modes. This is due to the use of the \glspl*{kde} and the choice of bandwidth $\sigma$.

\begin{figure*}
    \centering
    \def\x{0.125}
    \begin{subfigure}{0.05\textwidth}
    \centering
    \begin{tikzpicture}%
        \node () at (0, 17.5em) {\small{H}};
        \node () at (0, 11em) {\small{8}};
        \node () at (0, 4.5em) {\small{Box}};
        \node () at (0, 0em) {};
    \end{tikzpicture}%
    \end{subfigure}%
    \hfill%
    \begin{subfigure}{\x\textwidth}
    \includegraphics[width=\textwidth]{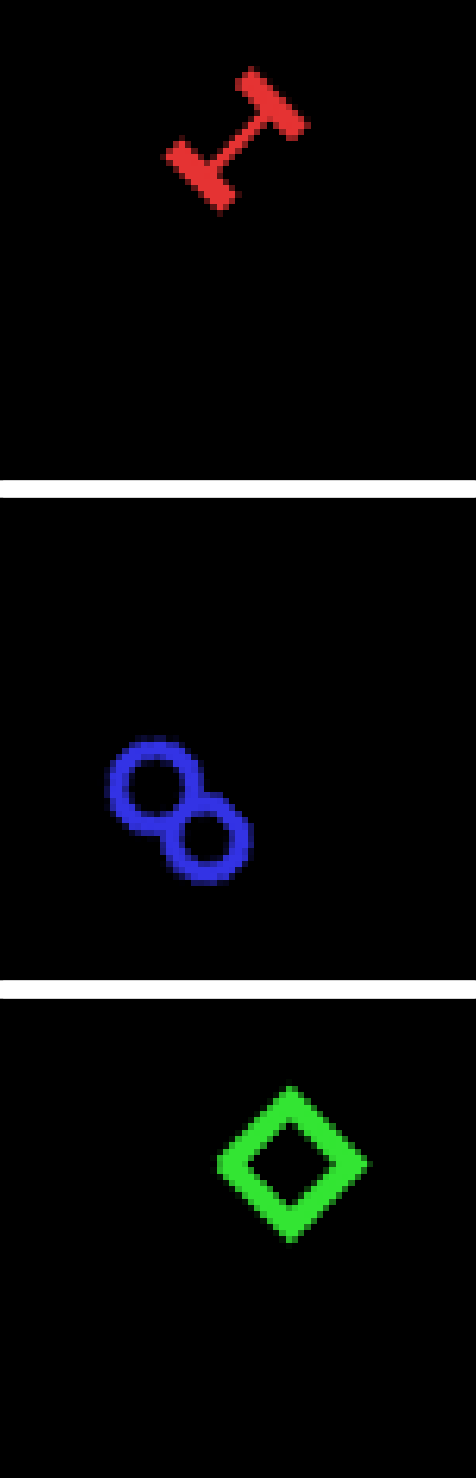}
    \caption{Problem}
    \end{subfigure}%
    \hfill%
    \begin{subfigure}{\x\textwidth}
    \includegraphics[width=\textwidth]{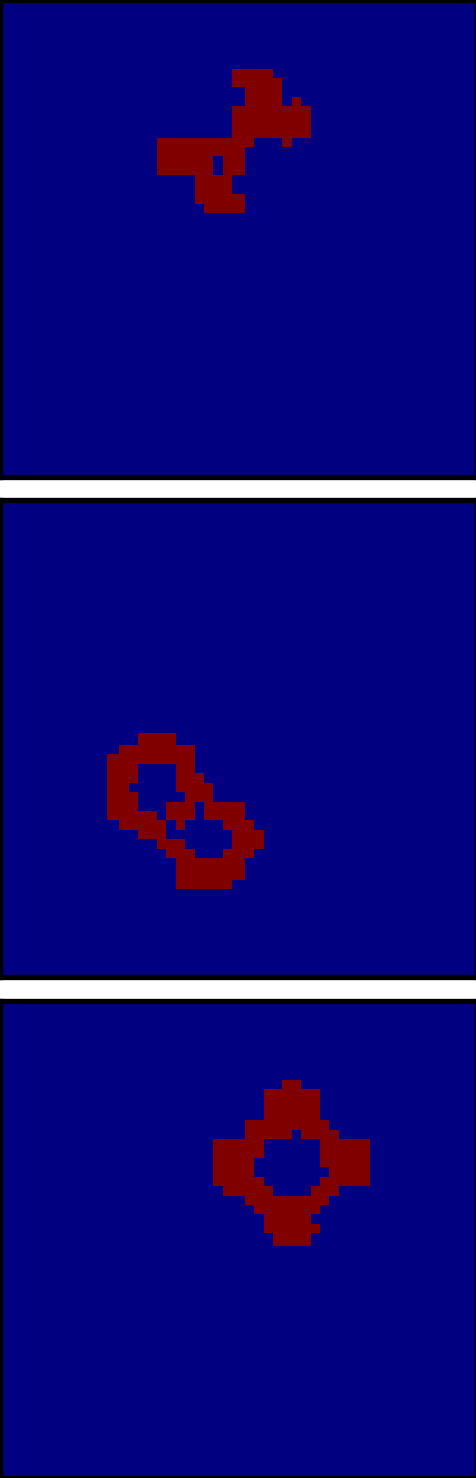}
    \caption{Truth}
    \end{subfigure}%
    \hfill%
    \begin{subfigure}{\x\textwidth}
    \includegraphics[width=\textwidth]{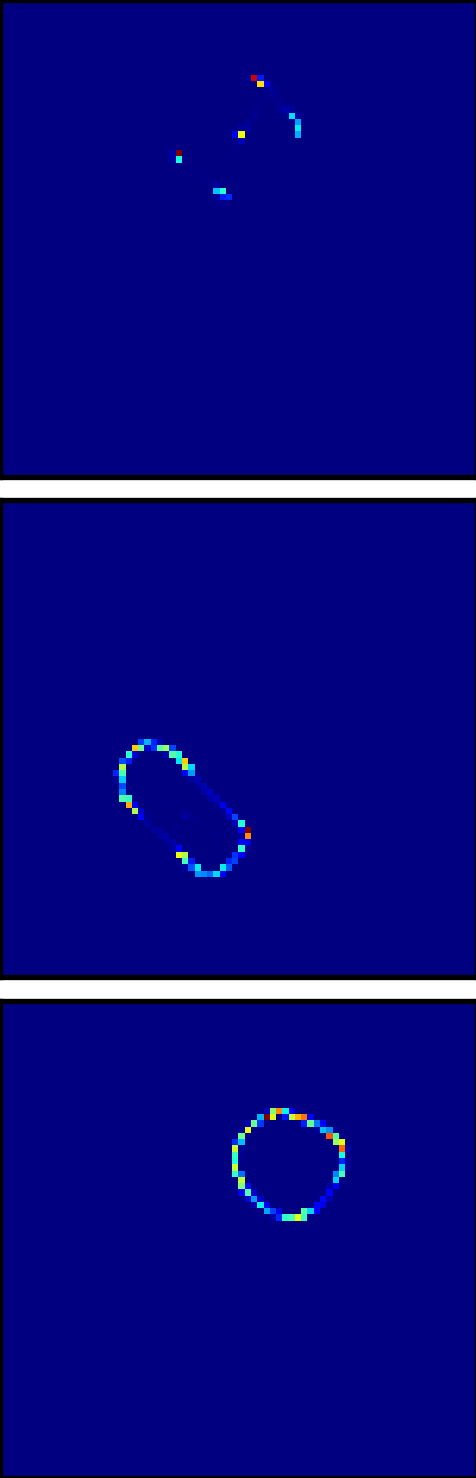}
    \caption{JS}
    \end{subfigure}%
    \hfill%
    \begin{subfigure}{\x\textwidth}
    \includegraphics[width=\textwidth]{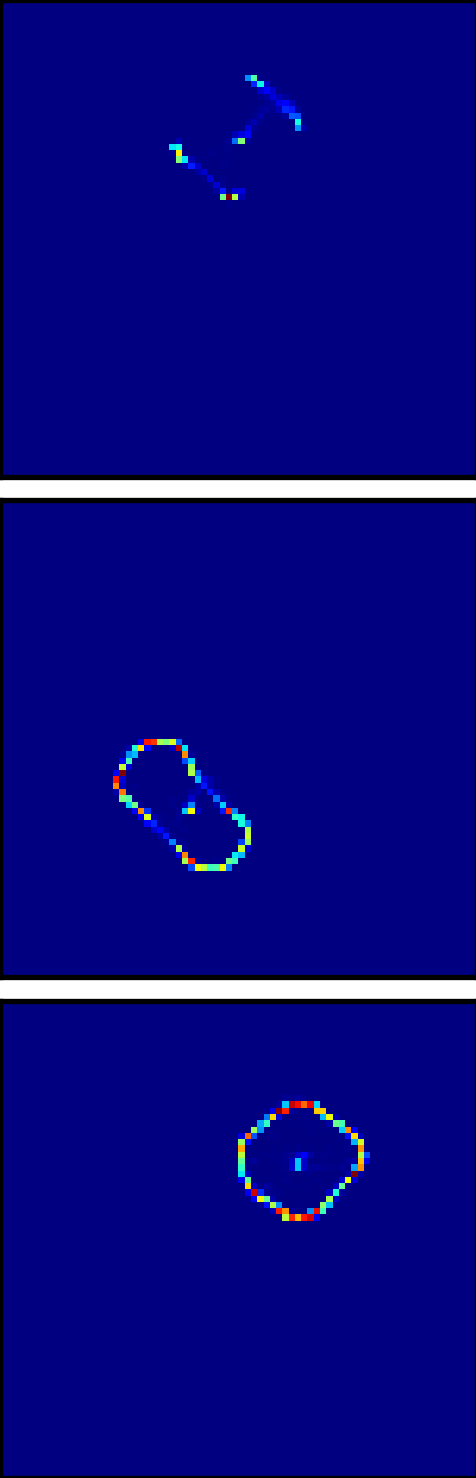}
    \caption{FKL}
    \end{subfigure}%
    \hfill%
    \begin{subfigure}{\x\textwidth}
    \includegraphics[width=\textwidth]{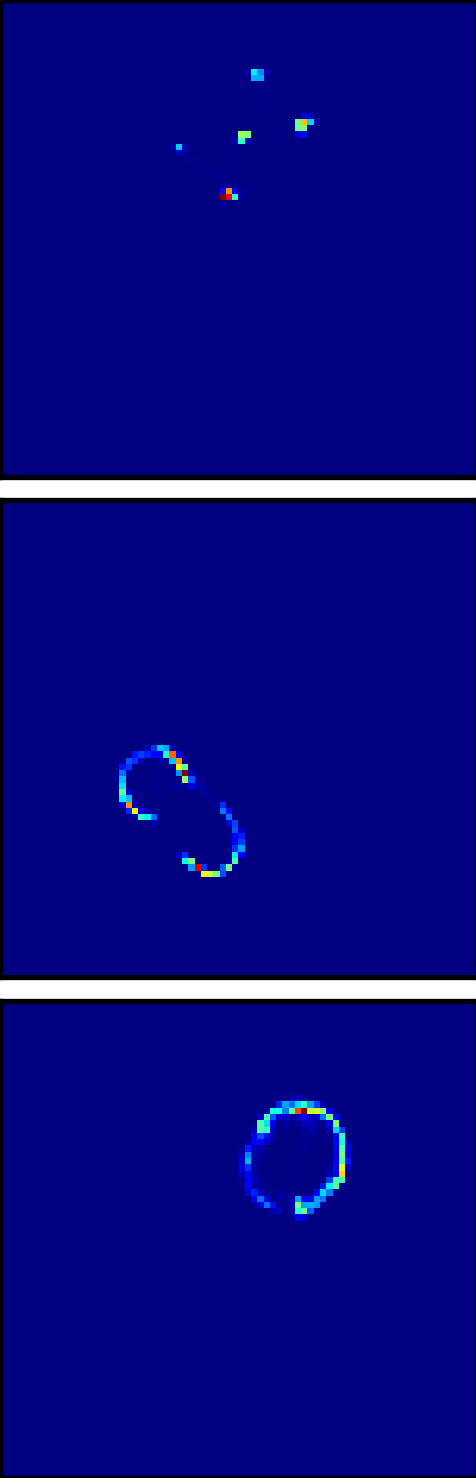}
    \caption{RKL}
    \end{subfigure}%
    \hfill%
    \begin{subfigure}{\x\textwidth}
    \includegraphics[width=\textwidth]{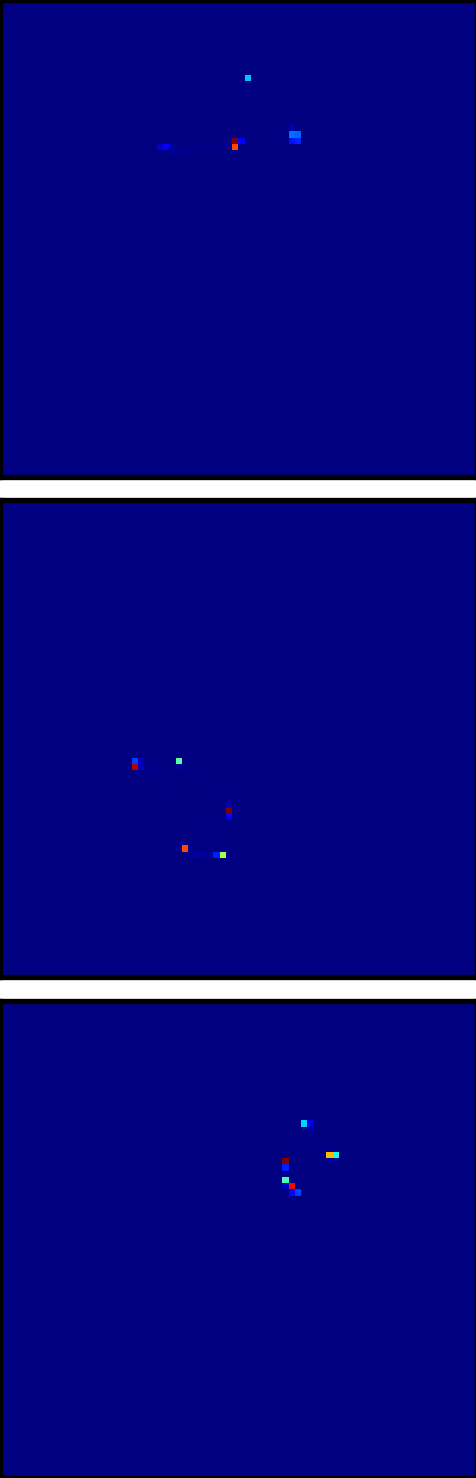}
    \caption{GAN}
    \end{subfigure}%
    \hfill%
    \begin{subfigure}{\x\textwidth}
    \includegraphics[width=\textwidth]{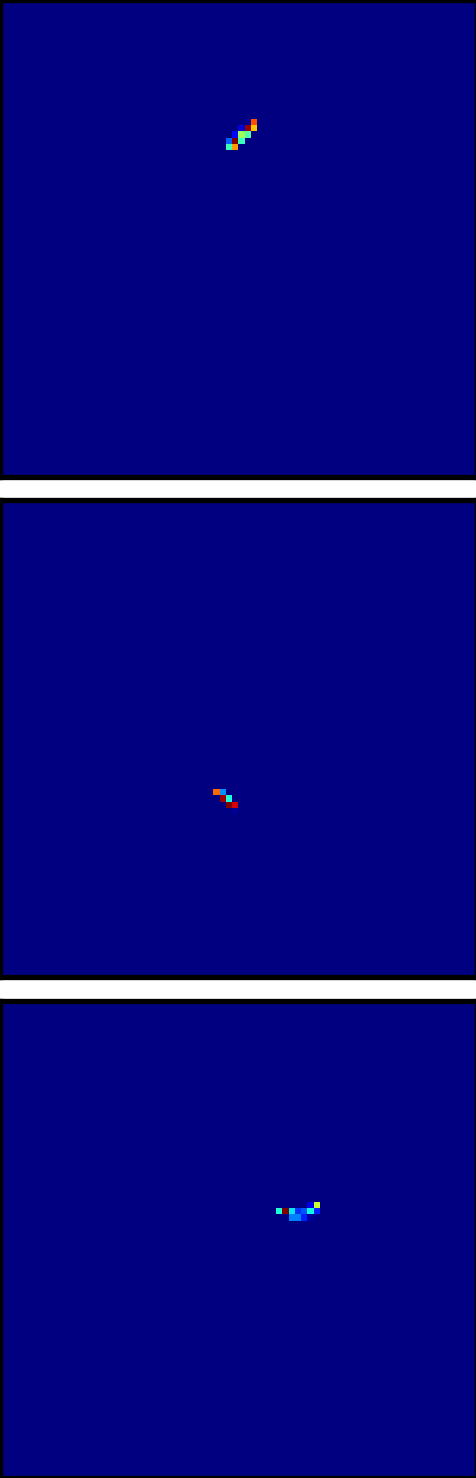}
    \caption{ME}
    \end{subfigure}\hfill%
    \caption{Qualitative comparison of the implemented algorithms, showing action heat-maps on three different states, with the last state never been observed during training.}
    \label{fig:qual}
\end{figure*}
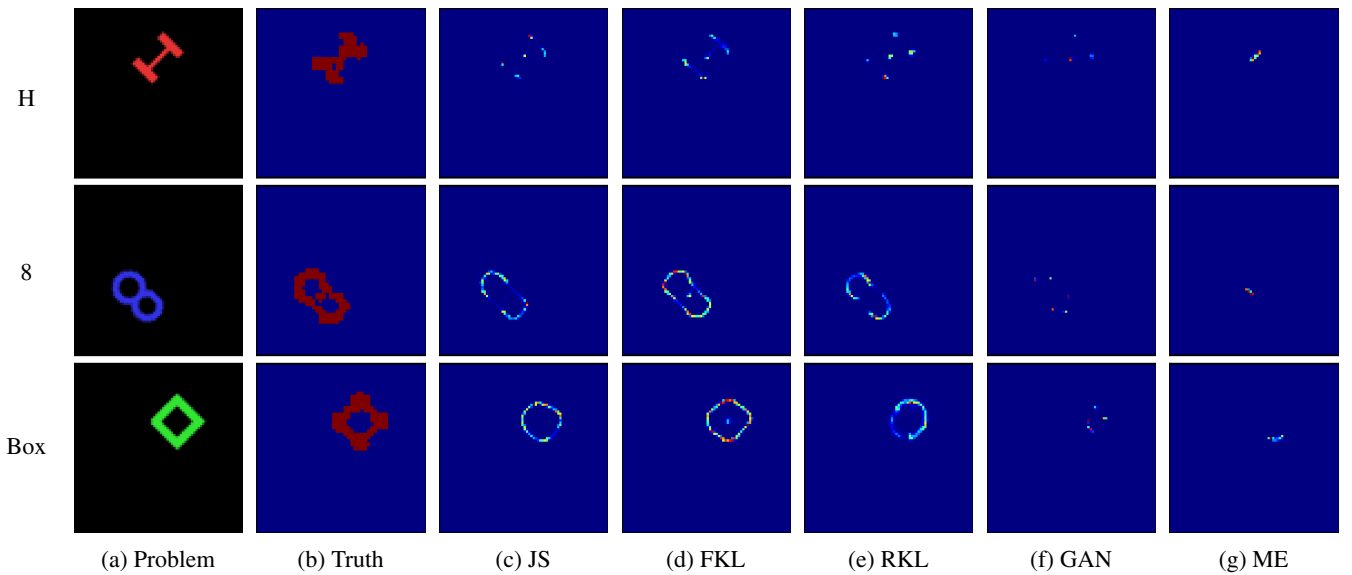

In the qualitative comparison in Figure~\ref{fig:qual}, the actor distributions of the different algorithms are shown for three different shapes. While the \textit{H} and \textit{8} shapes were trained on, the \textit{Box} shape has not been seen during training. The different subfigures show the action heat-maps of all implemented algorithms, showing only the x-y projections. The \textit{H}-row shows that \gls*{js} and \gls*{fkl} learned all five modes, with \gls*{js} having the fewest samples in the connecting area. Against the expectation from the illustrative examples, \gls*{rkl} also learned all modes. The most probable reason is that the actor learns to match the critic's distribution, changing simultaneously from a rough estimate of one feasibility region to the refined shape of individual modes. If the actor learns the entire distribution of the critic early on, when the critic learns to distinguish different modes, the actor's distribution has support in all modes and is thus trapped in each mode. The \gls*{gan} implementation shows four very unbalanced modes. Additionally, the modes are single points, which correspond to the automatically generated imitation examples, showing that the \gls*{gan} approach can only imitate but cannot find other feasible actions. The \gls*{me} implementation collapses in a single mode. The \textit{8}-row and the \textit{Box}-row show a similar pattern with the most pronounced spread of the action distributions in \gls*{js}, \gls*{fkl}, and \gls*{rkl} and mostly collapsed action regions in the other approaches. 

Each algorithm's accuracy and shares of modes on all shapes were evaluated to quantify the capability of generating actions in all disconnected sets of feasible actions. 1024 random states were generated for each shape that differed in pose, color, and geometry. For each state, 1024 actions were sampled from the different actors. The actions were then evaluated, and the mode of each action was recorded. The modes were then ranked and averaged over all the states of that shape by frequency. By averaging the ranks instead of the modes, the last rank shows the average ratio of the least frequent mode for each state. 

\begin{figure}
    \centering
    \begin{subfigure}{0.8\columnwidth}
    \includegraphics[width=\textwidth]{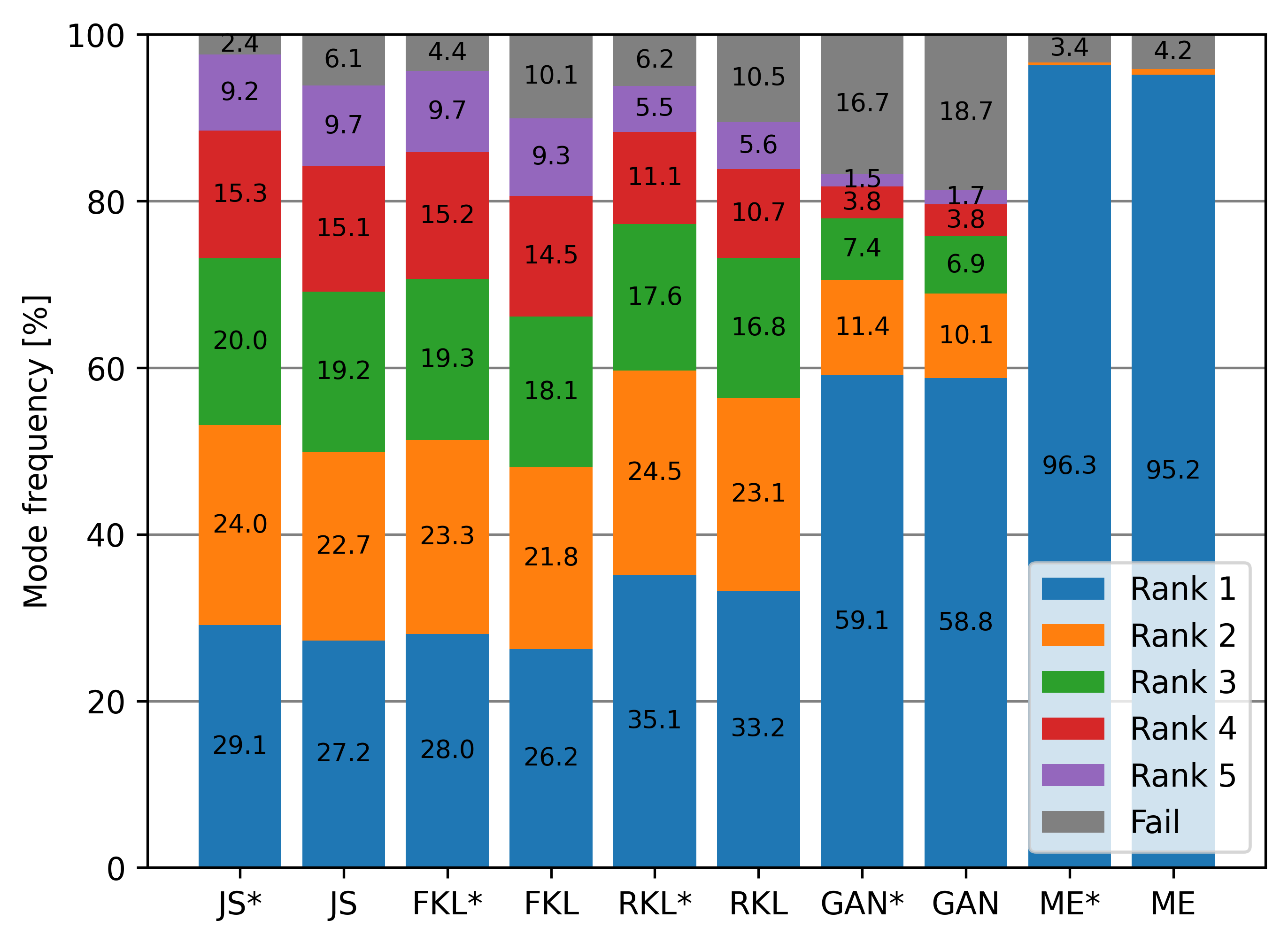}
    \caption{\textit{H} Shape}
    \end{subfigure}
    \begin{subfigure}{0.8\columnwidth}
    \includegraphics[width=\textwidth]{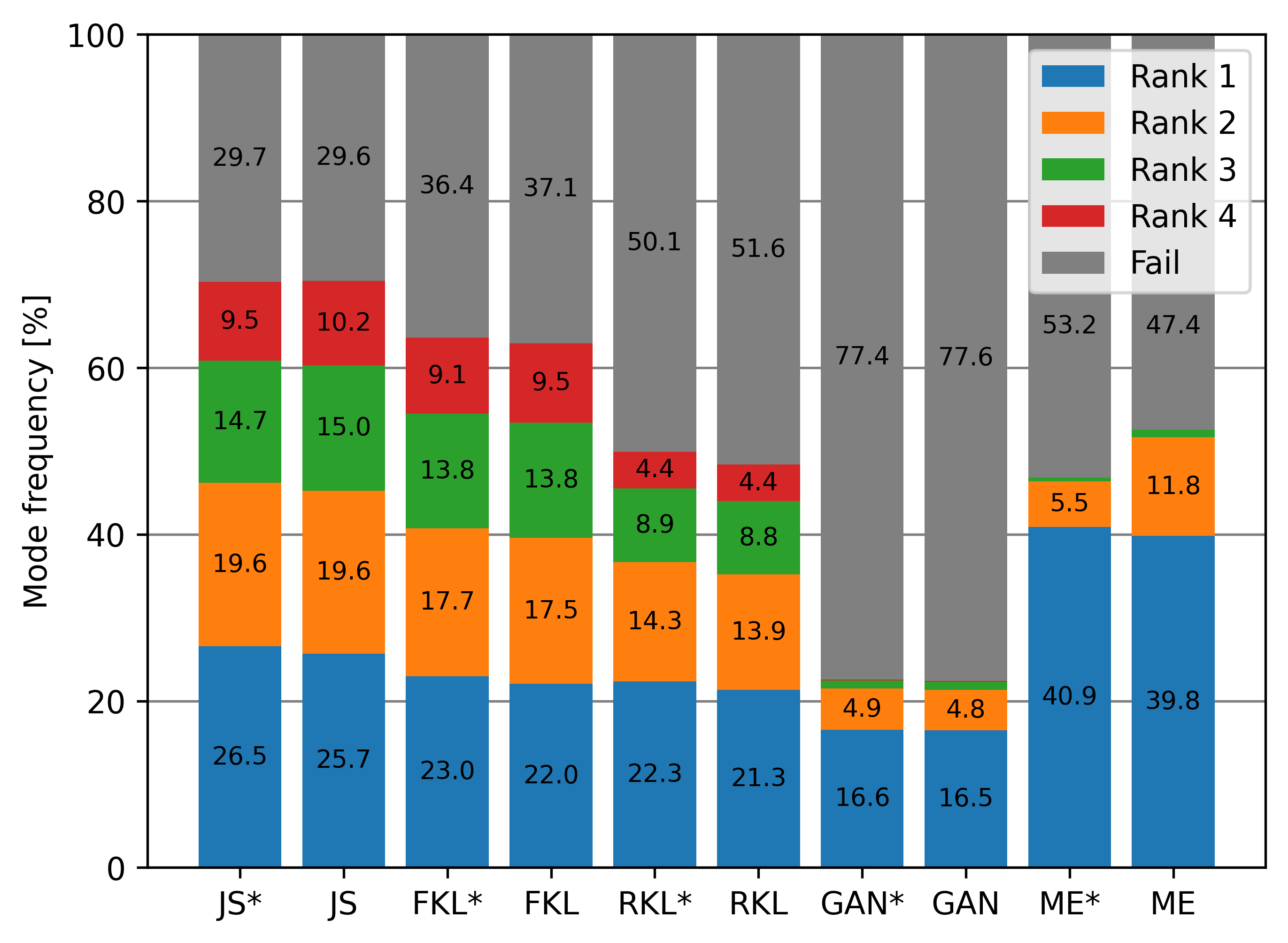}
    \caption{\textit{Box} Shape}
    \end{subfigure}
    \caption{Gripping rank comparison, with the ratio of picks for each ranked mode or failure in \%.}
    \label{fig:gripping_rank}
\end{figure}

Figure~\ref{fig:gripping_rank} shows the shares of each rank for the \textit{H} and \textit{Box} shapes for all the algorithms. This figure presents the multimodal capabilities of the proposed approaches. For the \textit{H} shape, \gls*{js} and \gls*{fkl} have the most balanced distribution over the grasping modes. The \gls*{gan} approach sometimes generates actions in all the modes but primarily focuses the actions in a primary mode. The \gls*{me} approach almost exclusively generates actions in one mode. The comparison on the \textit{Box} shape shows that the generalization capability of the \gls*{js} and \gls*{fkl} algorithms outperform the other approaches, which could indicate that explicitly learning all feasible actions improves generalization. The generalization capability of the \gls*{gan} implementation is significantly lower than the others, as seen on the \textit{Box} shape, indicating that that approach overfitted on the imitation examples.

To quantify the overall performance, Table~\ref{tab:data} shows the precision (feasible actions generated over total actions generated) for each shape and the last ranked mode for the \textit{H}, \textit{T}, and \textit{Box} shapes. The table shows that \gls*{me} has solid performance on all shapes trained on but has lower generalization performance and fails to find the different modes. The \gls*{gan} algorithm shows some actions in the last ranked modes, but it is significantly weaker overall. The best approach is \gls*{js} with the highest precision and similar shares in the last ranked mode as \gls*{fkl}. As discussed before, action optimization improves precision but reduces recall, slightly decreasing the least ranked mode for most approaches. The maximum deviations in the superscript show that all approaches learn reliably, with the \gls*{gan} having the highest performance deviations among runs.

\begin{table*}
\centering
\caption{\revise{Grasping score and mode comparison.}}
\setlength{\tabcolsep}{4pt}
\renewcommand*\arraystretch{1.2}
\begin{tabular}{cc||l|l|l|l|l|l||l|l|l|l}
& & JS* & JS & FKL* & FKL & RKL* & RKL & GAN* & GAN & ME* & ME 
\\ \hline
\multirow{6}{*}{\rotatebox{90}{Score}}
& H
& $\mathbf{97.6}^{0.0}$ & $93.9^{0.1}$ & $95.6^{0.5}$ & $89.9^{0.6}$ & $93.8^{1.6}$ & $89.5^{1.3}$ & $83.3^{5.9}$ & $81.3^{6.1}$ & $96.6^{0.3}$ & $95.8^{0.5}$  \\
& T
& $\mathbf{98.2}^{0.6}$ & $96.2^{0.6}$ & $97.1^{0.4}$ & $93.5^{0.1}$ & $96.1^{1.4}$ & $93.2^{1.3}$ & $84.8^{5.7}$ & $82.8^{5.3}$ & $96.7^{0.1}$ & $95.8^{0.7}$  \\
& 8
& $\mathbf{93.0}^{1.2}$ & $88.4^{1.4}$ & $89.5^{1.0}$ & $84.5^{0.7}$ & $87.3^{4.3}$ & $83.7^{4.6}$ & $58.9^{7.8}$ & $57.3^{7.9}$ & $86.6^{1.1}$ & $87.2^{2.6}$  \\
& Spoon
& $\mathbf{98.8}^{0.5}$ & $98.5^{0.5}$ & $97.4^{0.6}$ & $94.2^{1.4}$ & $98.2^{0.5}$ & $96.9^{0.9}$ & $86.5^{6.4}$ & $86.2^{6.9}$ & $96.7^{0.5}$ & $96.6^{0.7}$  \\
& Box
& $70.3^{4.3}$ & $\mathbf{70.4}^{4.0}$ & $63.6^{2.4}$ & $62.9^{2.1}$ & $49.9^{13.5}$ & $48.4^{12.7}$ & $22.6^{3.3}$ & $22.4^{4.0}$ & $46.8^{1.0}$ & $52.6^{16.0}$  \\
& Avg
& $\mathbf{91.6}^{0.8}$ & $89.5^{0.7}$ & $88.6^{0.3}$ & $85.0^{0.1}$ & $85.1^{3.9}$ & $82.3^{3.7}$ & $67.2^{5.2}$ & $66.0^{5.4}$ & $84.7^{0.1}$ & $85.6^{3.6}$  \\
\hline
\multirow{3}{*}{\rotatebox{90}{Mode}}
& H
& $9.2^{0.2}$ & $9.7^{0.3}$ & $\mathbf{9.7}^{0.8}$ & $9.3^{0.5}$ & $5.5^{0.3}$ & $5.6^{0.3}$ & $1.5^{2.2}$ & $1.7^{2.3}$ & $0.0^{0.0}$ & $0.0^{0.0}$  \\
& T
& $13.8^{0.8}$ & $14.4^{1.1}$ & $17.1^{0.6}$ & $\mathbf{17.3}^{0.2}$ & $9.1^{3.0}$ & $9.4^{3.0}$ & $2.0^{1.3}$ & $2.0^{1.3}$ & $0.0^{0.0}$ & $0.0^{0.0}$  \\
& Box
& $9.5^{1.2}$ & $\mathbf{10.2}^{0.9}$ & $9.1^{0.7}$ & $9.5^{0.7}$ & $4.4^{2.2}$ & $4.4^{2.0}$ & $0.1^{0.1}$ & $0.1^{0.1}$ & $0.0^{0.0}$ & $0.0^{0.0}$  \\\hline
\multicolumn{12}{p{390pt}}{\revise{Comparison on all shapes with the mean of the grasping success ratio in \% on top and the least ranked mode in \% on the bottom, with the maximum deviations over the three runs in superscript.}}
\end{tabular}
\label{tab:data}
\end{table*}

\subsection{Observation Variation Experiments}
\label{app:experiments}

\begin{figure}
    \centering
    \begin{subfigure}{0.8\columnwidth}
    \includegraphics[width=\textwidth]{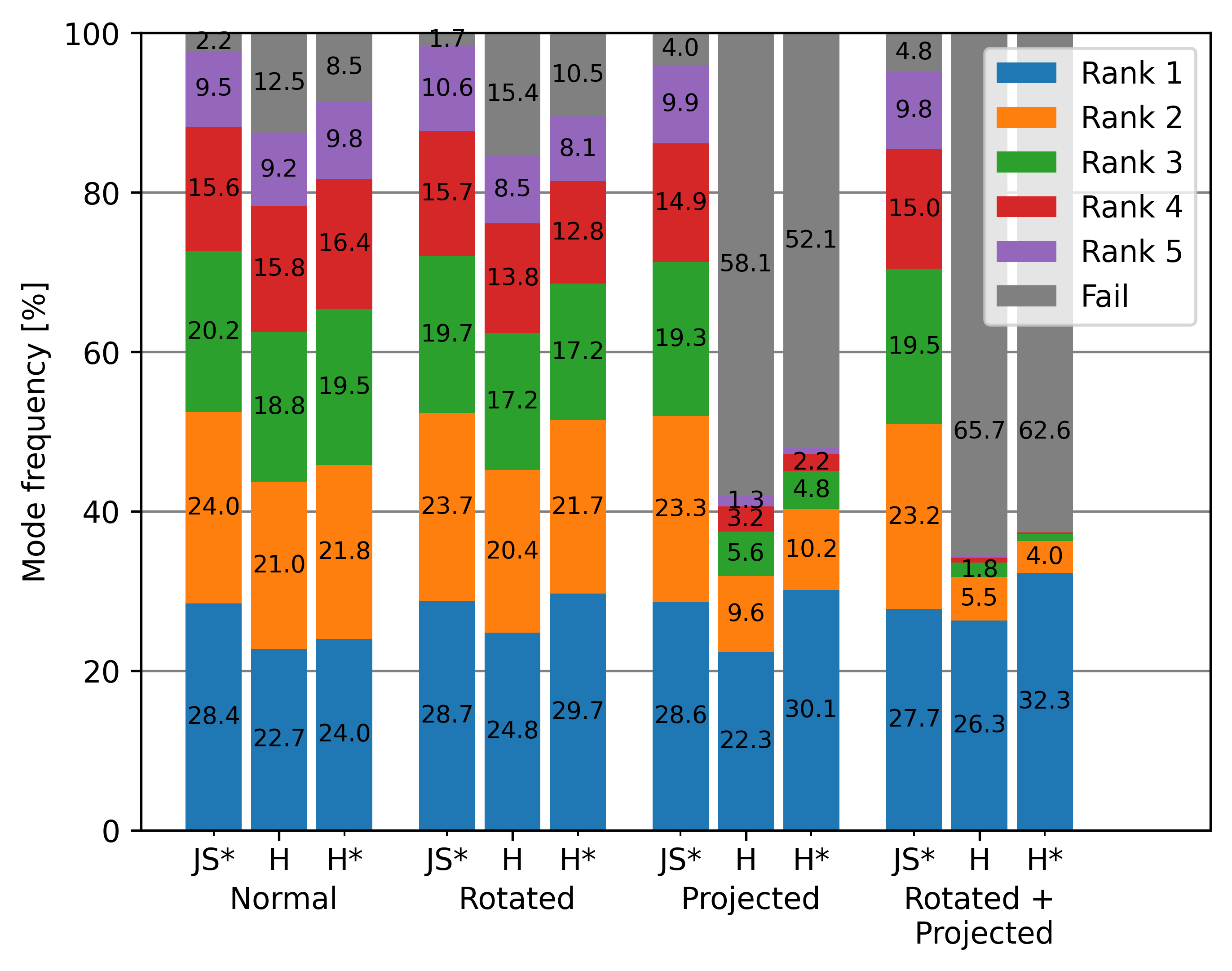}
    \caption{\textit{H} Shape}
    \end{subfigure}
    \begin{subfigure}{0.8\columnwidth}
    \includegraphics[width=\textwidth]{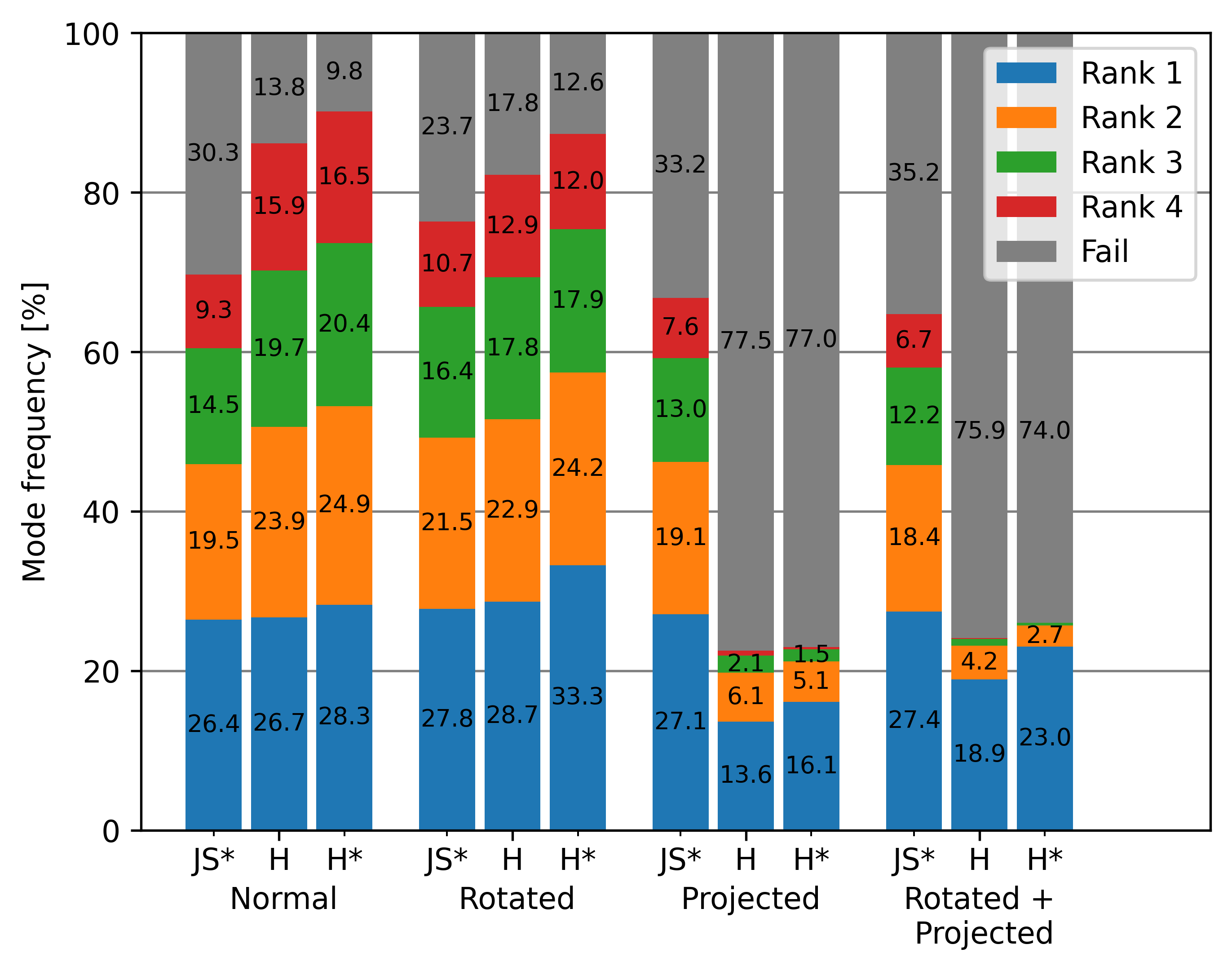}
    \caption{\textit{Box} Shape}
    \end{subfigure}
    \caption{Gripping rank comparison, with the ratio of picks for each ranked mode or failure in \%.}
    \label{fig:gripping_rank_distorted}
\end{figure}

\begin{table*}
\centering
\caption{\revise{Grasping score and mode comparison under perspective distortions.}}
\setlength{\tabcolsep}{4pt}
\renewcommand*\arraystretch{1.2}
\begin{tabular}{cc||c|c|c||c|c|c||c|c|c||c|c|c}
& & \multicolumn{3}{c||}{Normal}& \multicolumn{3}{c||}{Rotated}& \multicolumn{3}{c||}{Projected}& \multicolumn{3}{c}{Rotated + 
 Projected}\\ & & JS* & H & H* & JS* & H & H* & JS* & H & H* & JS* & H & H* 
\\ \hline
\multirow{6}{*}{\rotatebox{90}{Score}}
& H
& $\mathbf{97.8}$ & $87.5$& $91.5$& $\mathbf{98.3}$ & $84.6$& $89.5$& $\mathbf{96.0}$ & $41.9$& $47.9$& $\mathbf{95.2}$ & $34.3$& $37.4$ \\
& T
& $\mathbf{98.9}$ & $88.4$& $92.3$& $\mathbf{98.9}$ & $87.2$& $91.8$& $\mathbf{97.4}$ & $41.9$& $46.0$& $\mathbf{96.2}$ & $38.6$& $40.7$ \\
& 8
& $\mathbf{91.6}$ & $84.8$& $89.4$& $\mathbf{94.9}$ & $80.9$& $86.4$& $\mathbf{90.3}$ & $24.5$& $28.0$& $\mathbf{86.6}$ & $17.2$& $19.0$ \\
& Spoon
& $\mathbf{99.4}$ & $89.0$& $93.0$& $\mathbf{98.9}$ & $88.0$& $92.2$& $\mathbf{98.3}$ & $43.5$& $46.1$& $\mathbf{97.1}$ & $38.0$& $40.9$ \\
& Box
& $69.7$& $86.2$& $\mathbf{90.2}$ & $76.3$& $82.2$& $\mathbf{87.4}$ & $\mathbf{66.8}$ & $22.5$& $23.0$& $\mathbf{64.8}$ & $24.1$& $26.0$ \\
& Avg
& $\mathbf{91.5}$ & $87.2$& $91.3$& $\mathbf{93.5}$ & $84.6$& $89.4$& $\mathbf{89.8}$ & $34.9$& $38.2$& $\mathbf{88.0}$ & $30.5$& $32.8$ \\
\hline
\multirow{3}{*}{\rotatebox{90}{Mode}}
& H
& $9.5$& $9.2$& $\mathbf{9.8}$ & $\mathbf{10.6}$ & $8.5$& $8.1$& $\mathbf{9.9}$ & $1.3$& $0.7$& $\mathbf{9.8}$ & $0.2$& $0.0$ \\
& T
& $13.8$& $17.9$& $\mathbf{18.5}$ & $12.8$& $\mathbf{16.5}$ & $15.5$& $\mathbf{8.2}$ & $1.6$& $0.9$& $\mathbf{8.8}$ & $0.4$& $0.2$ \\
& Box
& $9.3$& $15.9$& $\mathbf{16.5}$ & $10.7$& $\mathbf{12.9}$ & $12.0$& $\mathbf{7.6}$ & $0.7$& $0.3$& $\mathbf{6.7}$ & $0.1$& $0.0$ \\
    \end{tabular}
    \label{tab:results_distorted}
\end{table*}
For each observation distortion, we trained one agent using the JS loss and one agent using the heat-map approach, each for $10^6$ training steps.
The results are shown in Figure~\ref{fig:gripping_rank_distorted} and Table~\ref{tab:results_distorted}, which compare the performance of the proposed Jensen-Shannon (JS) approach with the heat-map (H) approach.
As expected, the domain-specific heat-map approach performs well on the original problem. In that scenario, no scene understanding is required, and only local features need to be considered to estimate grasping success. Therefore, the approach is expected to generalize well to unseen shapes, as seen for the Box-Shape, since the grasping success depends only on gripper alignment. It only needs to learn to imitate the grasping success heuristic shown in Figure~\ref{fig:gripper_positions}. 

Rotating the observation does not seem to impact its performance. However, under projection and projection + rotation, the heat-map approach fails to learn to grasp reliably. Our proposed approach learns well under all distortions. In general, the performance of our proposed approach does not depend on the distortion as it does not explicitly use the spatial structure. Its design does not depend on the specifics of the experiment at all. It can, therefore, learn independently of the distortion applied as long as the object is still fully observable. 

\section{Discussion}
\label{sec:conclusion}

This \revise{paper} introduced the concept of action mapping, in which an optimization process can be learned sequentially by first learning feasibility and then learning the objective. In this \revise{paper}, we focused on the former part by learning to generate all feasible actions. We showed that by formulating a distribution matching problem and deriving a gradient estimator for general f-divergences, we train a feasibility policy that can function as a map between a latent space and the feasible action space. An illustrative example,\revise{ a robotic path planning example,} and experiments for robotic grasping show that our approach allows the feasibility policy to generate actions in all disconnected sets of feasible actions, a challenging task for state-of-the-art approaches. Enabling FKL and JS through our gradient estimator was instrumental.

\revise{Our experiments, detailed in Table}~\ref{tab:params}\revise{, reveal that training time varies significantly across different setups, with no clear correlation to increases in dimensionality. Surprisingly, the 2D system described in Section}~\ref{sec:illustrative} \revise{required more training time than the 4D system in Section}~\ref{sec:splines}\revise{. While our results do not show increased complexity with higher dimensions, we anticipate that scalability to higher-dimensional action spaces may still pose challenges. Nevertheless, adopting alternative non-parametric density estimators from existing literature could help mitigate these scalability concerns.}

Given the proposed method for training the feasibility policy from a feasibility model, the following steps will focus on action mapping. We will test it in reinforcement learning scenarios for which a feasibility model is known. A potential problem could be that the rough transition between disconnected sets of feasible actions makes deterministic objective policies more challenging. An added regularizing loss on smoothness could improve the transition, all be it by likely reducing accuracy. Further, the approach is very sensitive to the KDE bandwidth. We may need to adapt it throughout training, learn it, or derive a better estimate based on the Jacobian of the network.

\bibliographystyle{ieeetr}
\bibliography{bib}

\begin{thebibliography}{10}

\bibitem{bak2009system}
S.~Bak, D.~K. Chivukula, O.~Adekunle, M.~Sun, M.~Caccamo, and L.~Sha, ``The system-level simplex architecture for improved real-time embedded system safety,'' in {\em 2009 15th IEEE Real-Time and Embedded Technology and Applications Symposium}, pp.~99--107, IEEE, 2009.

\bibitem{bharadhwaj2021conservative}
H.~Bharadhwaj, A.~Kumar, N.~Rhinehart, S.~Levine, F.~Shkurti, and A.~Garg, ``Conservative safety critics for exploration,'' in {\em International Conference on Learning Representations}, 2021.

\bibitem{cheng2019end}
R.~Cheng, G.~Orosz, R.~M. Murray, and J.~W. Burdick, ``End-to-end safe reinforcement learning through barrier functions for safety-critical continuous control tasks,'' in {\em Proceedings of the AAAI conference on artificial intelligence}, vol.~33, pp.~3387--3395, 2019.

\bibitem{alshiekh2018safe}
M.~Alshiekh, R.~Bloem, R.~Ehlers, B.~K{\"o}nighofer, S.~Niekum, and U.~Topcu, ``Safe reinforcement learning via shielding,'' in {\em Proceedings of the AAAI conference on artificial intelligence}, vol.~32, 2018.

\bibitem{huang2022closer}
S.~Huang and S.~Onta{\~n}{\'o}n, ``A closer look at invalid action masking in policy gradient algorithms,'' in {\em The International FLAIRS Conference Proceedings}, vol.~35, 2022.

\bibitem{krasowski2020safe}
H.~Krasowski, X.~Wang, and M.~Althoff, ``Safe reinforcement learning for autonomous lane changing using set-based prediction,'' in {\em IEEE 23rd International Conference on Intelligent Transportation Systems (ITSC)}, 2020.

\bibitem{theile2023learning}
M.~Theile, H.~Bayerlein, M.~Caccamo, and A.~L. Sangiovanni-Vincentelli, ``Learning to recharge: Uav coverage path planning through deep reinforcement learning,'' 2023.

\bibitem{nazari2018reinforcement}
M.~Nazari, A.~Oroojlooy, L.~Snyder, and M.~Tak{\'a}c, ``Reinforcement learning for solving the vehicle routing problem,'' {\em Advances in Neural Information Processing Systems}, vol.~31, 2018.

\bibitem{garcia2015comprehensive}
J.~Garc{\i}a and F.~Fern{\'a}ndez, ``A comprehensive survey on safe reinforcement learning,'' {\em Journal of Machine Learning Research}, vol.~16, no.~1, pp.~1437--1480, 2015.

\bibitem{fisac2019general}
J.~F. Fisac, A.~K. Akametalu, M.~N. Zeilinger, S.~Kaynama, J.~Gillula, and C.~J. Tomlin, ``A general safety framework for learning-based control in uncertain robotic systems,'' {\em IEEE Transactions on Automatic Control}, vol.~64, no.~7, pp.~2737--2752, 2019.

\bibitem{li2018safe}
Z.~Li, U.~Kalabić, and T.~Chu, ``Safe reinforcement learning: Learning with supervision using a constraint-admissible set,'' in {\em 2018 Annual American Control Conference (ACC)}, pp.~6390--6395, 2018.

\bibitem{dalal2018safe}
G.~Dalal, K.~Dvijotham, M.~Vecerik, T.~Hester, C.~Paduraru, and Y.~Tassa, ``Safe exploration in continuous action spaces,'' {\em arXiv preprint arXiv:1801.08757}, 2018.

\bibitem{ames2019control}
A.~D. Ames, S.~Coogan, M.~Egerstedt, G.~Notomista, K.~Sreenath, and P.~Tabuada, ``Control barrier functions: Theory and applications,'' in {\em 2019 18th European control conference (ECC)}, pp.~3420--3431, IEEE, 2019.

\bibitem{sha2001using}
L.~Sha {\em et~al.}, ``Using simplicity to control complexity,''

\bibitem{zhong2021safe}
B.~Zhong, A.~Lavaei, H.~Cao, M.~Zamani, and M.~Caccamo, ``Safe-visor architecture for sandboxing (ai-based) unverified controllers in stochastic cyber--physical systems,'' {\em Nonlinear Analysis: Hybrid Systems}, vol.~43, p.~101110, 2021.

\bibitem{hastings1970}
W.~K. Hastings, ``{Monte Carlo sampling methods using Markov chains and their applications},'' {\em Biometrika}, vol.~57, pp.~97--109, 04 1970.

\bibitem{gelfand1990}
A.~E. Gelfand and A.~F. Smith, ``Sampling-based approaches to calculating marginal densities,'' {\em Journal of the American statistical association}, vol.~85, no.~410, pp.~398--409, 1990.

\bibitem{kruschke2015}
J.~K. Kruschke, ``Chapter 5 - bayes' rule,'' in {\em Doing Bayesian Data Analysis (Second Edition)} (J.~K. Kruschke, ed.), pp.~99--120, Boston: Academic Press, second edition~ed., 2015.

\bibitem{Jordan1998}
M.~Jordan, Z.~Ghahramani, T.~Jaakkola, and L.~Saul, ``An introduction to variational methods for graphical models,'' {\em Machine Learning}, vol.~37, pp.~183--233, 01 1999.

\bibitem{WainwrightJordan2008}
M.~Wainwright and M.~Jordan, ``Graphical models, exponential families, and variational inference,'' {\em Foundations and Trends in Machine Learning}, vol.~1, pp.~1--305, 01 2008.

\bibitem{nowozin2016}
S.~Nowozin, B.~Cseke, and R.~Tomioka, ``f-gan: Training generative neural samplers using variational divergence minimization,'' in {\em Advances in Neural Information Processing Systems} (D.~Lee, M.~Sugiyama, U.~Luxburg, I.~Guyon, and R.~Garnett, eds.), vol.~29, Curran Associates, Inc., 2016.

\bibitem{Hu2018}
T.~Hu, Z.~Chen, H.~Sun, J.~Bai, M.~Ye, and G.~Cheng, ``Stein neural sampler,'' {\em ArXiv}, vol.~abs/1810.03545, 2018.

\bibitem{Rezende2015}
D.~J. Rezende and S.~Mohamed, ``Variational inference with normalizing flows,'' in {\em ICML}, 2015.

\bibitem{Tabak2013}
E.~G. Tabak and C.~V. Turner, ``A family of nonparametric density estimation algorithms,'' {\em Communications on Pure and Applied Mathematics}, vol.~66, 2013.

\bibitem{Tabak2010}
E.~G. Tabak and E.~Vanden-Eijnden, ``Density estimation by dual ascent of the log-likelihood,'' {\em Communications in Mathematical Sciences}, vol.~8, pp.~217--233, 2010.

\bibitem{kong2020expressive}
Z.~Kong and K.~Chaudhuri, ``The expressive power of a class of normalizing flow models,'' in {\em International conference on artificial intelligence and statistics}, pp.~3599--3609, PMLR, 2020.

\bibitem{pmlr-v139-koehler21a}
F.~Koehler, V.~Mehta, and A.~Risteski, ``Representational aspects of depth and conditioning in normalizing flows,'' in {\em Proceedings of the 38th International Conference on Machine Learning} (M.~Meila and T.~Zhang, eds.), vol.~139 of {\em Proceedings of Machine Learning Research}, pp.~5628--5636, PMLR, 18--24 Jul 2021.

\bibitem{kalashnikov2018scalable}
D.~Kalashnikov, A.~Irpan, P.~Pastor, J.~Ibarz, A.~Herzog, E.~Jang, D.~Quillen, E.~Holly, M.~Kalakrishnan, V.~Vanhoucke, {\em et~al.}, ``Scalable deep reinforcement learning for vision-based robotic manipulation,'' in {\em Conference on Robot Learning}, pp.~651--673, PMLR, 2018.

\bibitem{kleeberger2020survey}
K.~Kleeberger, R.~Bormann, W.~Kraus, and M.~F. Huber, ``A survey on learning-based robotic grasping,'' {\em Current Robotics Reports}, vol.~1, no.~4, pp.~239--249, 2020.

\bibitem{kumra2020antipodal}
S.~Kumra, S.~Joshi, and F.~Sahin, ``Antipodal robotic grasping using generative residual convolutional neural network,'' in {\em 2020 IEEE/RSJ International Conference on Intelligent Robots and Systems (IROS)}, pp.~9626--9633, IEEE, 2020.

\bibitem{morrison2020learning}
D.~Morrison, P.~Corke, and J.~Leitner, ``Learning robust, real-time, reactive robotic grasping,'' {\em The International journal of robotics research}, vol.~39, no.~2-3, pp.~183--201, 2020.

\bibitem{zeng2020tossingbot}
A.~Zeng, S.~Song, J.~Lee, A.~Rodriguez, and T.~Funkhouser, ``Tossingbot: Learning to throw arbitrary objects with residual physics,'' {\em IEEE Transactions on Robotics}, vol.~36, no.~4, pp.~1307--1319, 2020.

\bibitem{liese2006}
F.~Liese and I.~Vajda, ``On divergences and informations in statistics and information theory,'' {\em IEEE Transactions on Information Theory}, vol.~52, no.~10, pp.~4394--4412, 2006.

\bibitem{lecuyer1995sufficient}
P.~L'Ecuyer, ``On the interchange of derivative and expectation for likelihood ratio derivative estimators,'' {\em Management Science}, vol.~41, no.~4, pp.~738--748, 1995.

\bibitem{Kleijnen1996optimization}
J.~Kleijnen and R.~Rubinstein, ``{Optimization and Sensitivity Analysis of Computer Simulation Models by the Score Function Method},'' Other publications TiSEM 958c9b9a-544f-48f3-a3d1-c, Tilburg University, School of Economics and Management, 1996.

\bibitem{Murphy2012}
K.~P. Murphy, {\em Machine Learning: A Probabilistic Perspective}.
\newblock The MIT Press, 2012.

\bibitem{He2016}
K.~He, X.~Zhang, S.~Ren, and J.~Sun, ``Deep residual learning for image recognition,'' {\em 2016 IEEE Conference on Computer Vision and Pattern Recognition (CVPR)}, pp.~770--778, 2016.

\bibitem{haarnoja2018soft}
T.~Haarnoja, A.~Zhou, P.~Abbeel, and S.~Levine, ``Soft actor-critic: Off-policy maximum entropy deep reinforcement learning with a stochastic actor,'' in {\em International conference on machine learning}, pp.~1861--1870, PMLR, 2018.

\bibitem{mirza2014}
M.~Mirza and S.~Osindero, ``Conditional generative adversarial nets,'' {\em ArXiv}, vol.~abs/1411.1784, 2014.

\bibitem{tensorflow2015-whitepaper}
M.~Abadi, A.~Agarwal, P.~Barham, E.~Brevdo, Z.~Chen, C.~Citro, G.~S. Corrado, A.~Davis, J.~Dean, M.~Devin, S.~Ghemawat, I.~Goodfellow, A.~Harp, G.~Irving, M.~Isard, Y.~Jia, R.~Jozefowicz, L.~Kaiser, M.~Kudlur, J.~Levenberg, D.~Man\'{e}, R.~Monga, S.~Moore, D.~Murray, C.~Olah, M.~Schuster, J.~Shlens, B.~Steiner, I.~Sutskever, K.~Talwar, P.~Tucker, V.~Vanhoucke, V.~Vasudevan, F.~Vi\'{e}gas, O.~Vinyals, P.~Warden, M.~Wattenberg, M.~Wicke, Y.~Yu, and X.~Zheng, ``{TensorFlow}: Large-scale machine learning on heterogeneous systems,'' 2015.
\newblock Software available from tensorflow.org.

\end{thebibliography}

\appendices

\begin{IEEEbiography}[{\includegraphics[width=1in,height=1.25in,clip,keepaspectratio]{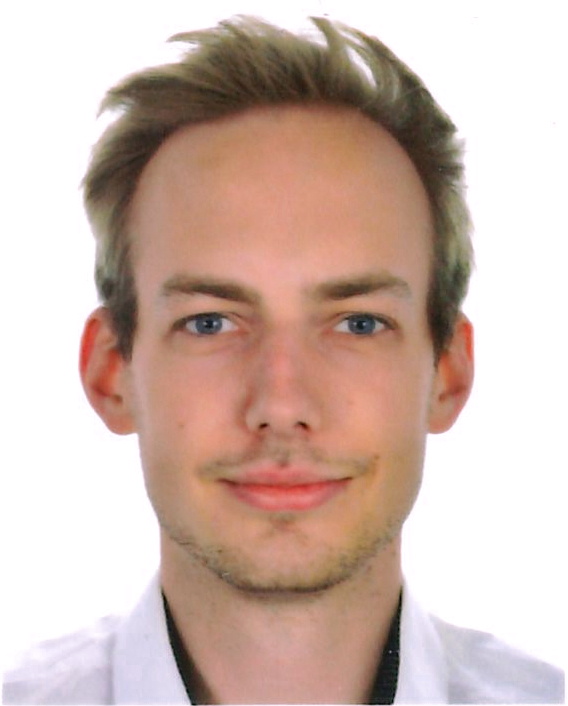}}]{Mirco Theile} 
received the M.Sc. degree in electrical engineering and information technology from Technical University of Munich, Germany, in 2018, where he is currently pursuing a Ph.D. degree. Currently, he is also a visiting researcher at the University of California in Berkeley, USA. His research interests extend to reinforcement learning in applications of cyber-physical systems, including UAVs, robotics, and real-time systems.
\end{IEEEbiography}

\begin{IEEEbiography}[{\includegraphics[width=1in,height=1.25in,clip,keepaspectratio]{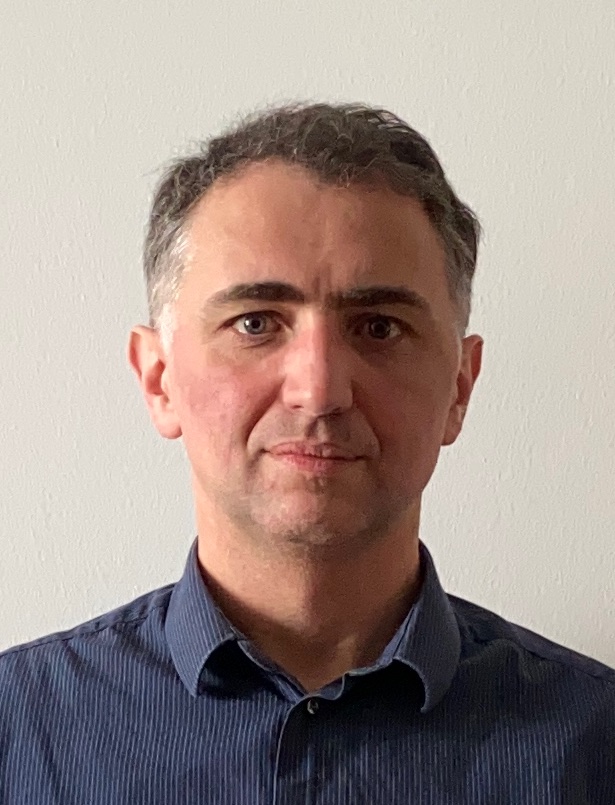}}]{Daniele Bernardini}
received his M.Sc. degree in Theoretical Physics at the University of Florence in 1997. After graduation, he spent 2 more years as a researcher at the Ludwig Maximilians University, Munich before transitioning to the industry, where he gained more than 20 years of experience in software development and data science. In 2021 he joined the Technical University of Munich as research group leader where he focuses on advancing perception for robotic manipulation. Since 2021 he is a co-founder and CEO of Cognivix, a startup specializing in automation solutions for industries requiring high variability and low volume production.
\end{IEEEbiography}

\begin{IEEEbiography}[{\includegraphics[width=1in,height=1.25in,clip,keepaspectratio]{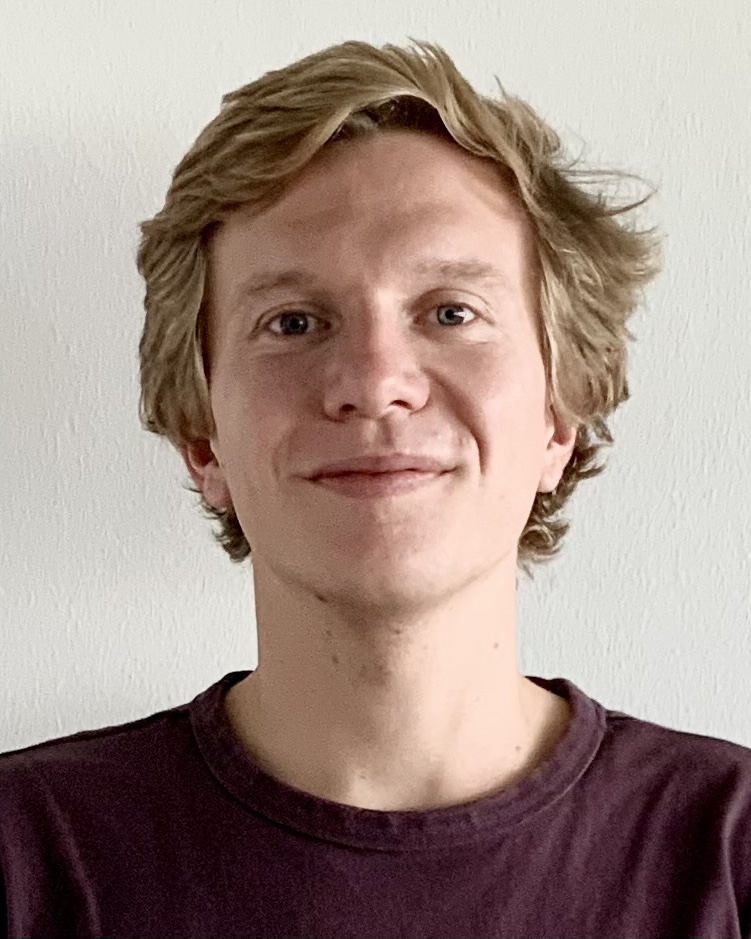}}]{Raphael Trumpp}
graduated with a M.Sc. degree in mechanical engineering from the Technical University of Munich in 2021, where he is currently pursuing a Ph.D. in informatics. His research focuses on machine learning, especially combining deep reinforcement learning with classical control methods. He is interested in applying these to interactive multi-agent scenarios like autonomous racing and robotics.
\end{IEEEbiography}

\begin{IEEEbiography}[{\includegraphics[width=1in,height=1.25in,clip,keepaspectratio]{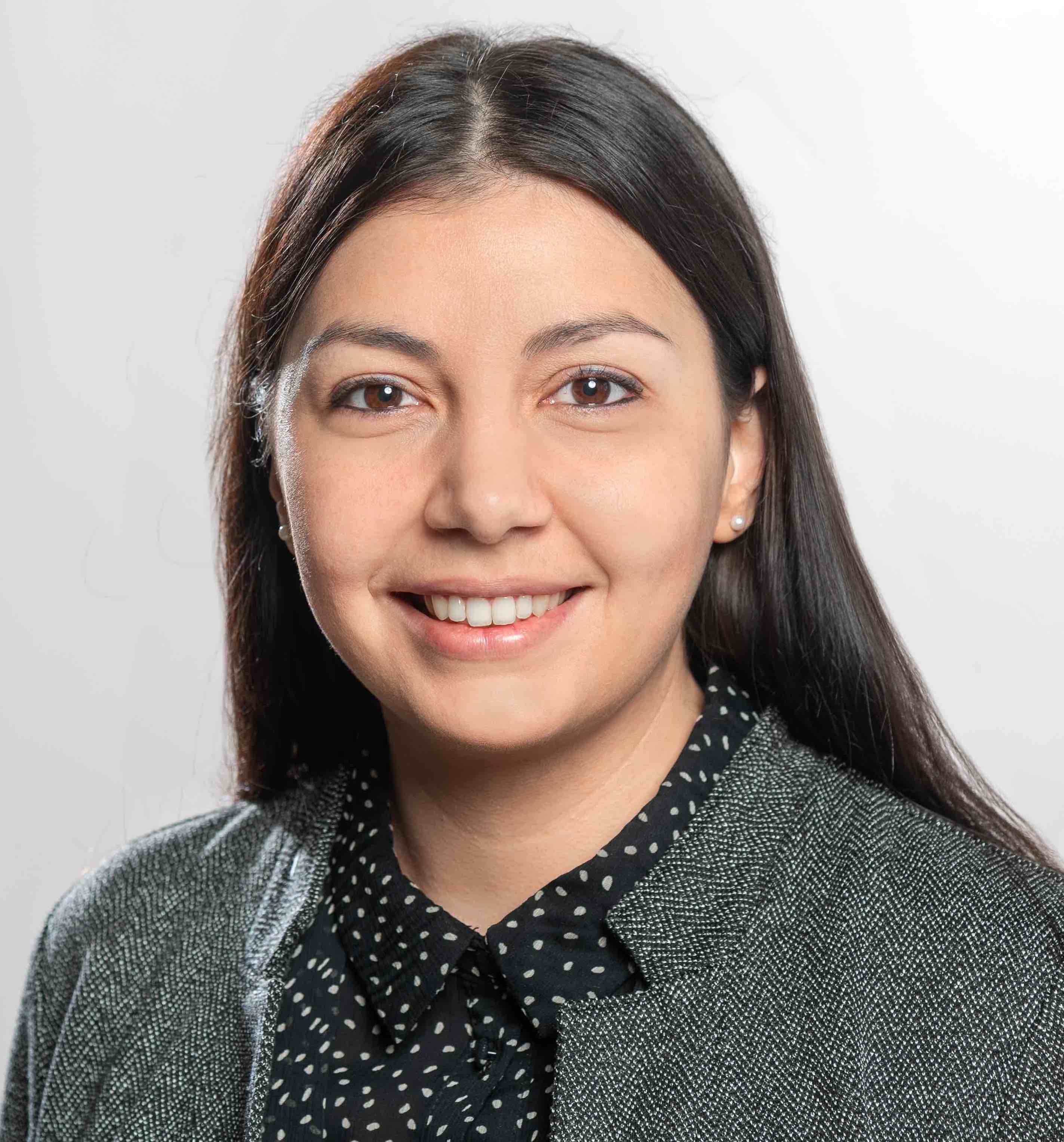}}]{Cristina Piazza} 
received a B.Sc. in Biomedical Engineering, a M.S. in Automation and Robotics Engineering and a PhD degree in Robotics (summa cum laude, 2019) from the University of Pisa (Italy). She subsequently moved to Chicago (USA) where she worked as a postdoctoral researcher at Northwestern University. Since 2020, Prof. Piazza is tenure track assistant professor at Technical University of Munich
\end{IEEEbiography}

\begin{IEEEbiography}[{\includegraphics[width=1in,height=1.25in,clip,keepaspectratio]{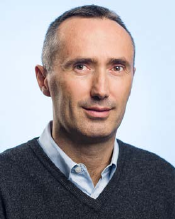}}]{Marco Caccamo}
earned his Ph.D. in computer engineering from Scuola Superiore Sant’Anna (Italy) in 2002. Shortly after graduation, he joined University of Illinois at Urbana-Champaign as assistant professor in Computer Science and was promoted to full professor in 2014. Since 2018, Prof. Caccamo has been appointed to the chair of Cyber-Physical Systems in Production Engineering at Technical University of Munich, Germany. In 2003, he was awarded an NSF CAREER Award. He is a recipient of the Alexander von Humboldt Professorship and he is IEEE Fellow.
\end{IEEEbiography}

\begin{IEEEbiography}[{\includegraphics[width=1in,height=1.25in,clip,keepaspectratio]{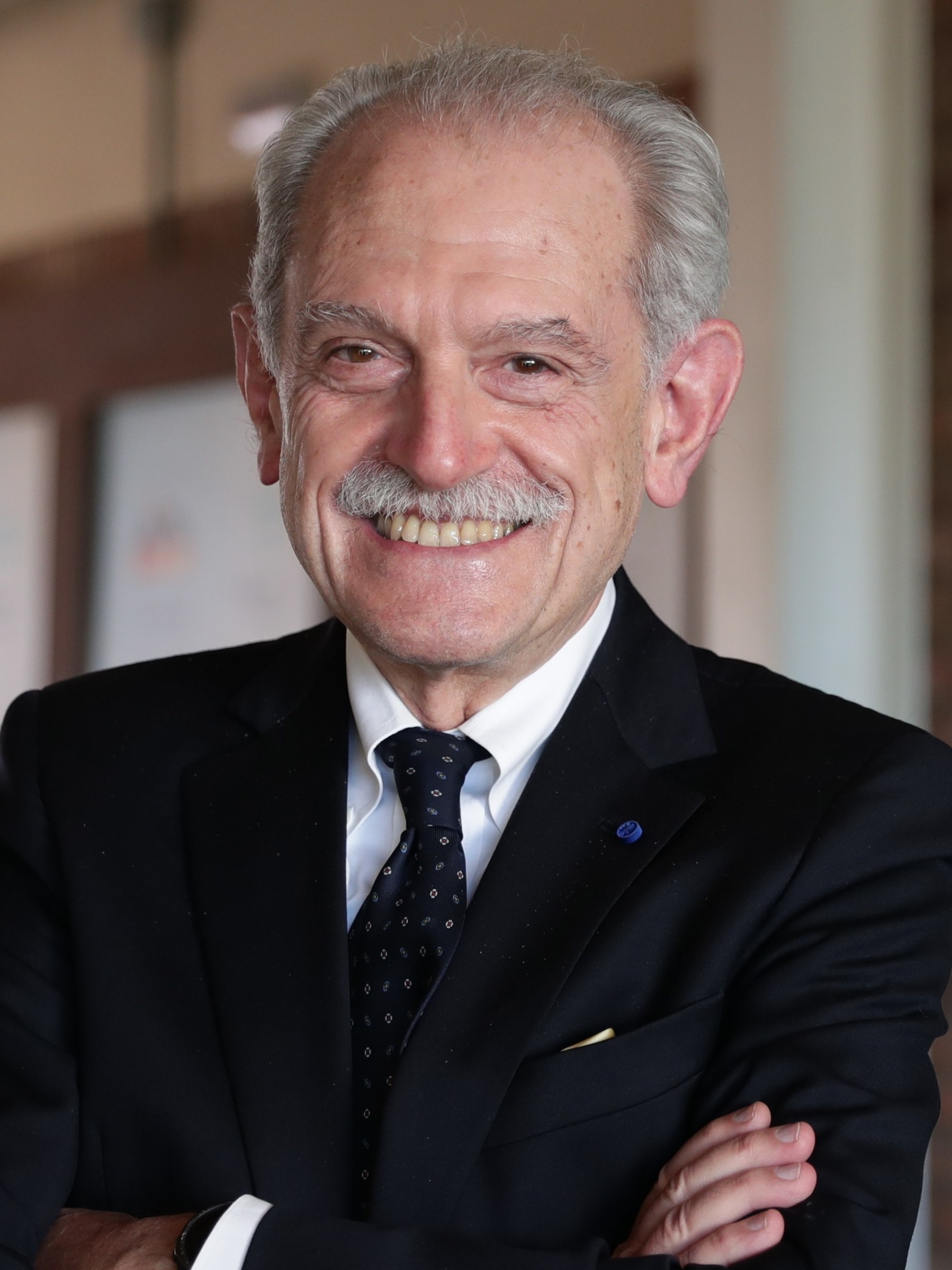}}]{Alberto L. Sangiovanni-Vincentelli} 
is the Edgar L. and Harold H. Buttner Chair of Electrical Engineering and Computer Sciences at the University of California, Berkeley. He was a co-founder of Cadence and Synopsys, the two leading companies in the area of Electronic Design Automation. He is currently a Board member of 8 companies, including Cadence, and Chairman of the Board of Quantum Motion, Innatera, Phoelex, e4Life and Phononic Vibes. He is the recipient of several academic honors, and research awards including the IEEE/RSE Wolfson James Clerk Maxwell Medal “for groundbreaking contributions that have had an exceptional impact on the development of electronics and electrical engineering or related fields”, the BBVA Frontiers of Knowledge Award in the Information and Communication Technologies category, the Kaufmann Award for seminal contributions to EDA, the IEEE Darlington Award, the EDAA lifetime Achievement Award, and four Honorary Doctorates from University of Aalborg, KTH, AGH and University of Rome, Tor Vergata. He is an author of over 1000 papers, 17 books and 3 patents in the area of design tools and methodologies, large scale systems, embedded systems, hybrid systems, and AI.
\end{IEEEbiography}
\vfill
\EOD

\end{document}